\documentclass[10pt,twocolumn,letterpaper]{article}

\usepackage{cvpr}              %
\usepackage{amsmath,amsfonts}

\usepackage{multirow}
\usepackage{svg}
\usepackage{bm}
\usepackage{adjustbox}
\usepackage{multirow}
\usepackage{cuted} 
\usepackage{pifont}%
\usepackage{graphicx}
\usepackage{booktabs}

\newcommand{\cmark}{\ding{51}}%
\newcommand{\xmark}{\ding{55}}%

\usepackage[ruled,vlined]{algorithm2e}
\usepackage{algorithmicx}
\usepackage{algpseudocode}
\usepackage{xr}

\usepackage{placeins}
\usepackage{pgfplots}
\pgfplotsset{compat=1.18}
\usepackage{pgfplotstable}
\usepgfplotslibrary{statistics}

\usepackage{caption}
\usepackage{subcaption}

\definecolor{cvprblue}{rgb}{0.21,0.49,0.74}
\usepackage[pagebackref,breaklinks,colorlinks,citecolor=cvprblue]{hyperref}

\def\paperID{10551} %
\def\confName{CVPR}
\def\confYear{2025}
\def\method{\textit{LD}}
\def\methodname{Latent Drifting}

\title{Latent Drifting in Diffusion Models for Counterfactual Medical Image Synthesis}

\author{Yousef Yeganeh$^{1,2}$\thanks{,$^\ddagger$ Equal Contribution}~, Azade Farshad$^{1,2}$\footnotemark[1] \thanks{Project Lead}~, Ioannis Charisiadis$^{1,\ddagger}$, Marta Hasny$^{1,\ddagger}$ , Martin Hartenberger$^{1,\ddagger}$ , \\Björn Ommer$^{2,3}$, Nassir Navab$^{1,2}$, Ehsan Adeli$^4$\footnotemark[2]\\
\\ $^1$Technical University of Munich, Munich, Germany \\
$^2$Munich Center for Machine Learning, Munich, Germany \\
$^3$CompVis @ LMU Munich, Munich, Germany\\
$^4$Stanford University, Stanford, CA, USA\\
{\tt\small \{y.yeganeh,azade.farshad,ge83cid,marta.hasny,martin.hartenberger,nassir.navab\}@tum.de},\\{\tt\small b.ommer@lmu.de},~{\tt\small eadeli@stanford.edu}
}

\begin{document}
\maketitle
\begin{strip}
\begin{center}
\vspace{-2.4cm}
    \centering
    \resizebox{\linewidth}{!}{
\begin{tabular}[b]{c@{\hspace{0.25em}}c@{\hspace{0.25em}}c@{\hspace{0.5em}}c@{\hspace{0.25em}}c@{\hspace{0.25em}}c@{\hspace{0.5em}}c@{\hspace{0.25em}}c@{\hspace{0.25em}}c}
    \multicolumn{3}{c}{Alzheimer's Disease $\rightarrow$ Healthy} & \multicolumn{3}{c}{Young $\rightarrow$ Old} &  \multicolumn{3}{c}{Disease: Pneumonia}\\
            \includegraphics[width=0.11\linewidth]{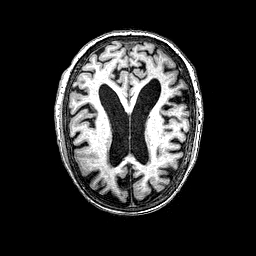}  &  \includegraphics[width=0.11\linewidth]{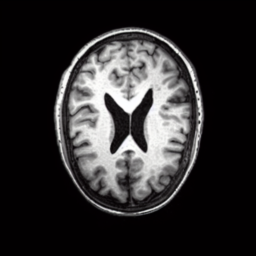} & \frame{\includegraphics[width=0.11\linewidth]{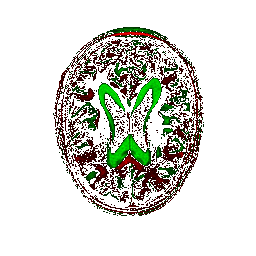}}  & 
            \includegraphics[width=0.11\linewidth]{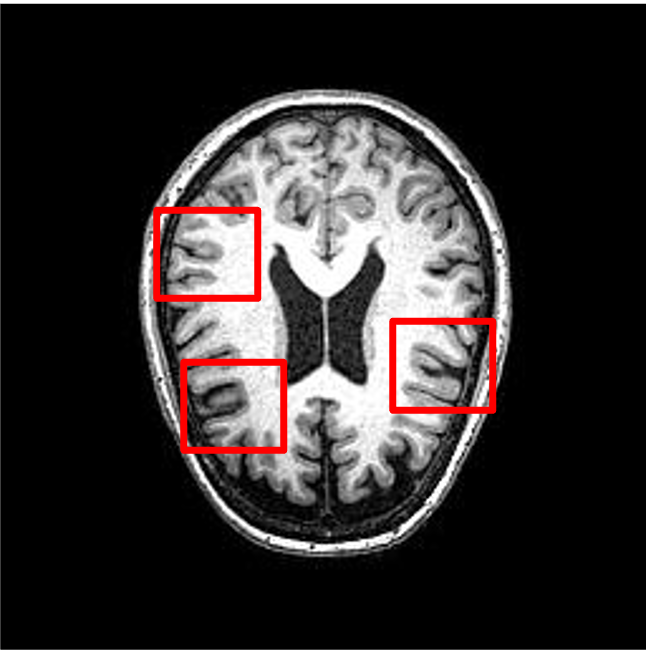} & \includegraphics[width=0.11\linewidth]{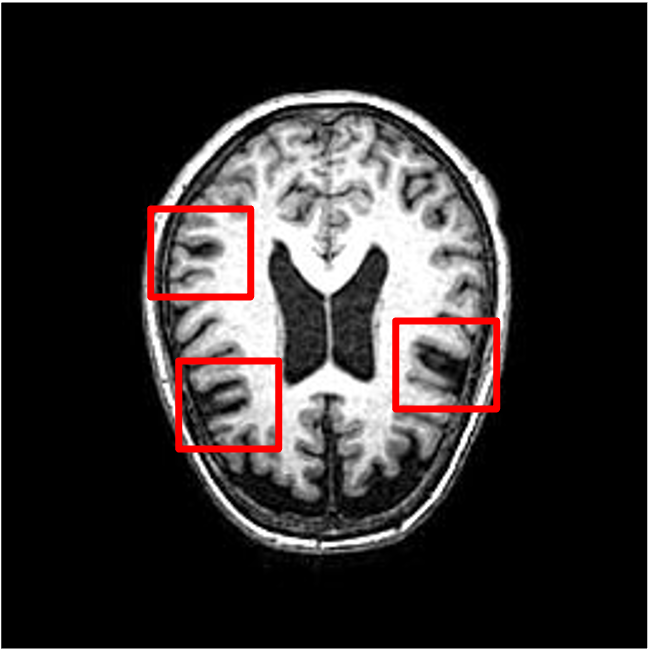} &   \includegraphics[width=0.11\linewidth]{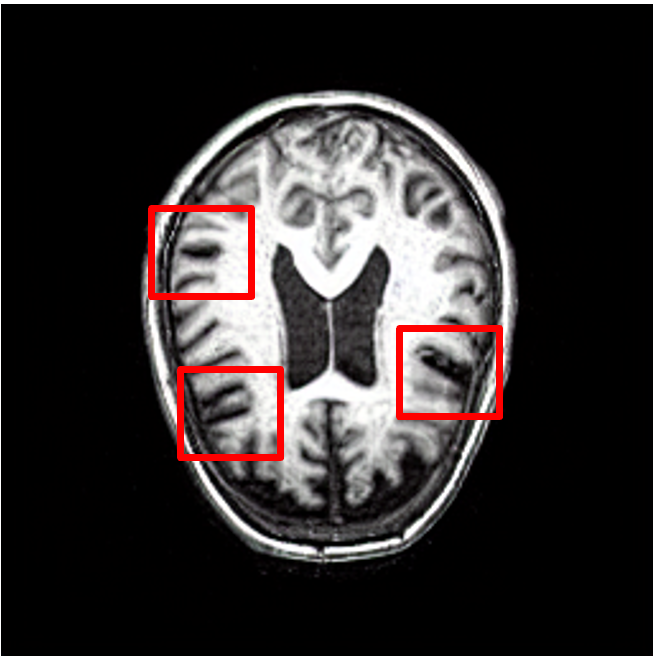} &    \includegraphics[width=0.11\linewidth]{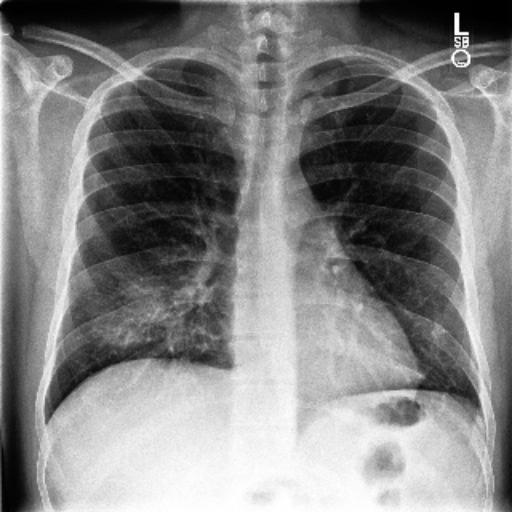}  & \includegraphics[width=0.11\linewidth]{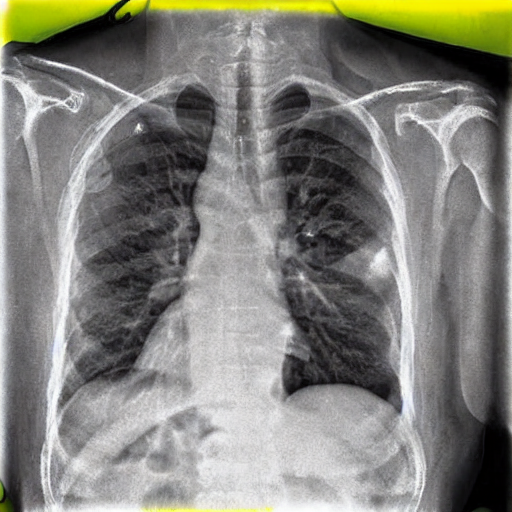}&      \includegraphics[width=0.11\linewidth]{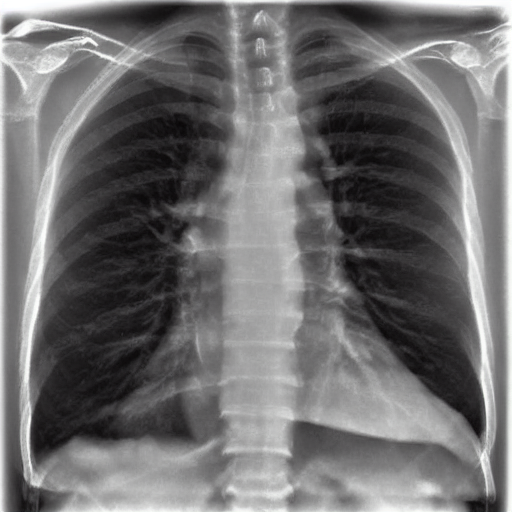} 
            \\
          \multicolumn{3}{c}{Healthy $\rightarrow$ Alzheimer's Disease } & \multicolumn{3}{c}{Young $\rightarrow$ Old}  & \multicolumn{3}{c}{Disease: Pleural Effusion} \\
          \includegraphics[width=0.11\linewidth]{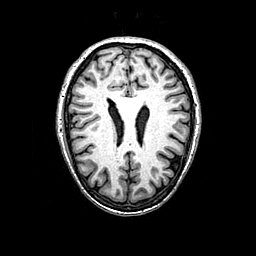}  &  \includegraphics[width=0.11\linewidth]{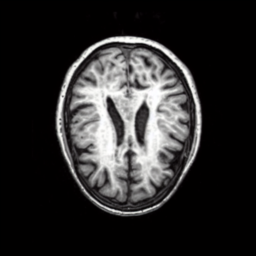} & \frame{\includegraphics[width=0.11\linewidth]{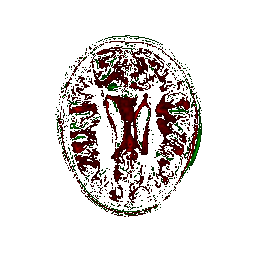}}  & \includegraphics[width=0.11\linewidth]{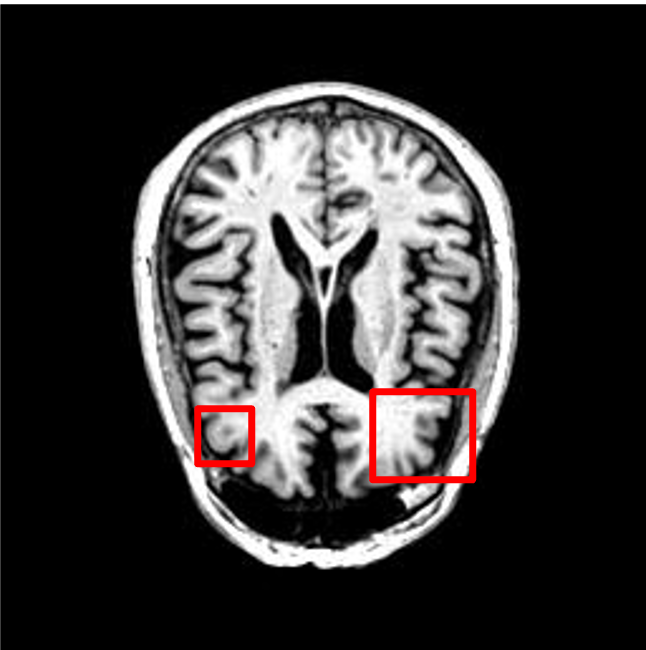} & \includegraphics[width=0.11\linewidth]{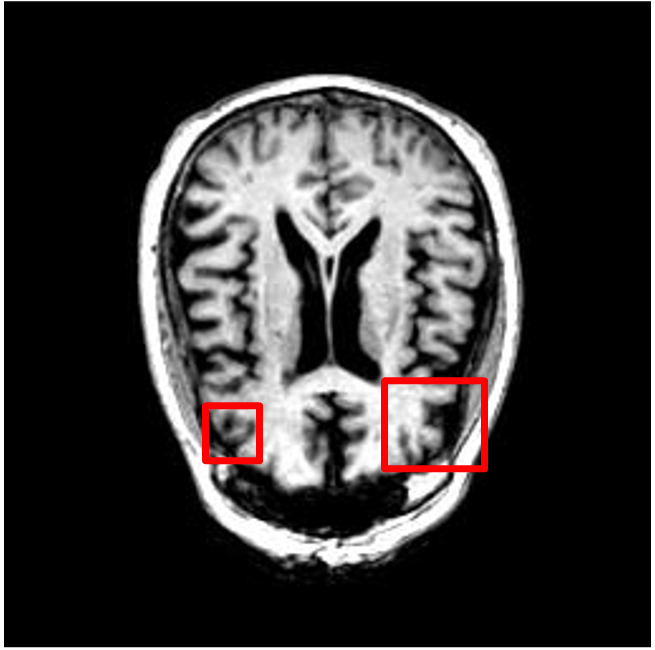} &\includegraphics[width=0.11\linewidth]{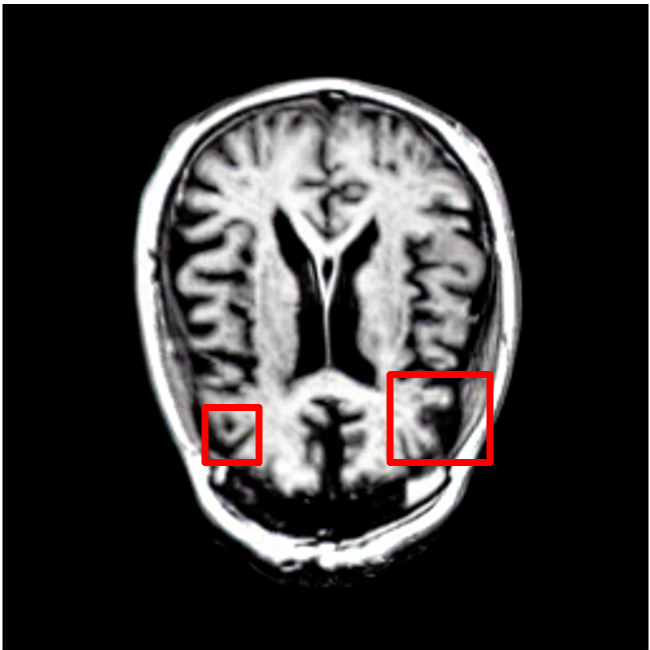}&              \includegraphics[width=0.1089\linewidth]{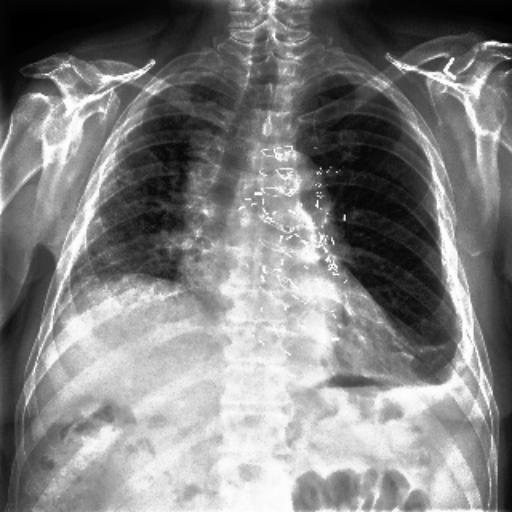} & \includegraphics[width=0.1089\linewidth]{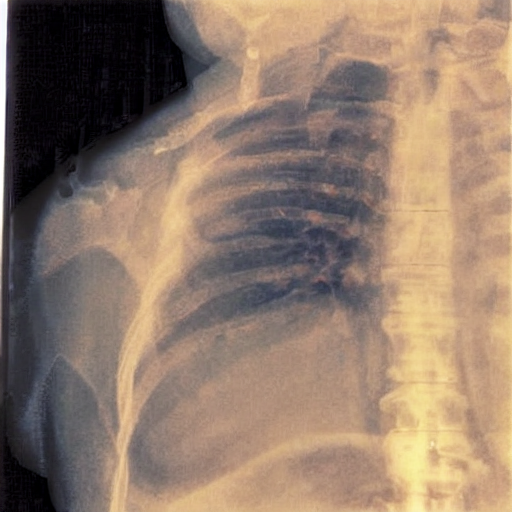}& \includegraphics[width=0.1089\linewidth]{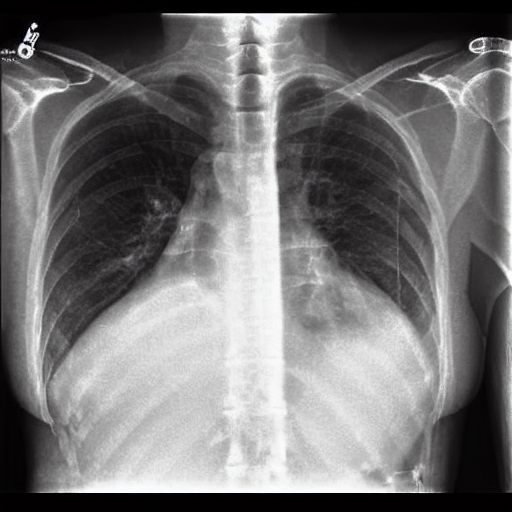} \\
          Source & LD (Ours) & Difference & Source & Target & LD (Ours) & Real Sample & w/o LD & LD (Ours)
          \\
        \end{tabular}
        }
\captionof{figure}{Medical Image Generation and Manipulation using \textbf{\method{}}.  Left to right: (1) \textit{(text, image)-to-image} without ground-truth (GT) pairs {\raisebox{0.7ex}{\colorbox{red}{}}: Removal, \raisebox{0.7ex}{\colorbox{green}{}}: Addition} , (2) Aging: \textit{(text, image)-to-image} with GT pairs, (3) \textit{text-to-image} without GT pairs.     \label{fig:interpret}} %
\end{center}
\end{strip}

\begin{abstract}
{\def\thefootnote{}\footnotetext{Project page: \href{https://latentdrifting.github.io/}{https://latentdrifting.github.io/}}}
Scaling by training on large datasets has been shown to enhance the quality and fidelity of image generation and manipulation with diffusion models; however, such large datasets are not always accessible in medical imaging due to cost and privacy issues, which contradicts one of the main applications of such models to produce synthetic samples where real data is scarce. Also, fine-tuning on pre-trained general models has been a challenge due to the distribution shift between the medical domain and the pre-trained models. Here, we propose Latent Drift (\method{}) for diffusion models that can be adopted for any fine-tuning method to mitigate the issues faced by the distribution shift or employed in inference time as a condition. \methodname{} enables diffusion models to be conditioned for medical images fitted for the complex task of counterfactual image generation, which is crucial to investigate how parameters such as gender, age, and adding or removing diseases in a patient would alter the medical images. We evaluate our method on three public longitudinal benchmark datasets of brain MRI and chest X-rays for counterfactual image generation. Our results demonstrate significant performance gains in various scenarios when combined with different fine-tuning schemes.

\end{abstract}

\section{Introduction}
\label{sec:intro}

In recent years, high-resolution image generation models such as the latent diffusion model (LDM) \cite{rombach2022high} have gained popularity for their ability to generate photorealistic images from prompts. These models are trained on large datasets such as LAION-5B \cite{schuhmann2022laion}, consisting of billions of primarily natural images with corresponding captions. Leveraging such models for medical image generation can be impactful where publicly available data, especially annotated, are scarce. Several factors contribute to the scarcity of medical data, including privacy concerns, the cost of collecting data from clinically ill patients, and the rarity of some diseases \cite{tajbakhsh2020embracing}. Additionally, conditional diffusion models hold valuable potential for counterfactual tasks like disease progression, aging, gender alteration, etc.
\newlength{\mytablewidth}
\setlength{\mytablewidth}{\linewidth}
\addtolength{\mytablewidth}{-1em} %
\newlength{\mywidth}
\setlength{\mywidth}{0.2\mytablewidth}

\begin{figure}[tb]
    \centering
    \begin{tabular}{c@{\hspace{0.2em}}c@{\hspace{0.2em}}c@{\hspace{0.2em}}c@{\hspace{0.2em}}c}
         \multicolumn{1}{c}{$\delta =-0.1$} & \multicolumn{1}{c}{-0.05} & \multicolumn{1}{c}{0} & \multicolumn{1}{c}{0.05} & \multicolumn{1}{c}{0.1} \\ 
\includegraphics[width=\mywidth]{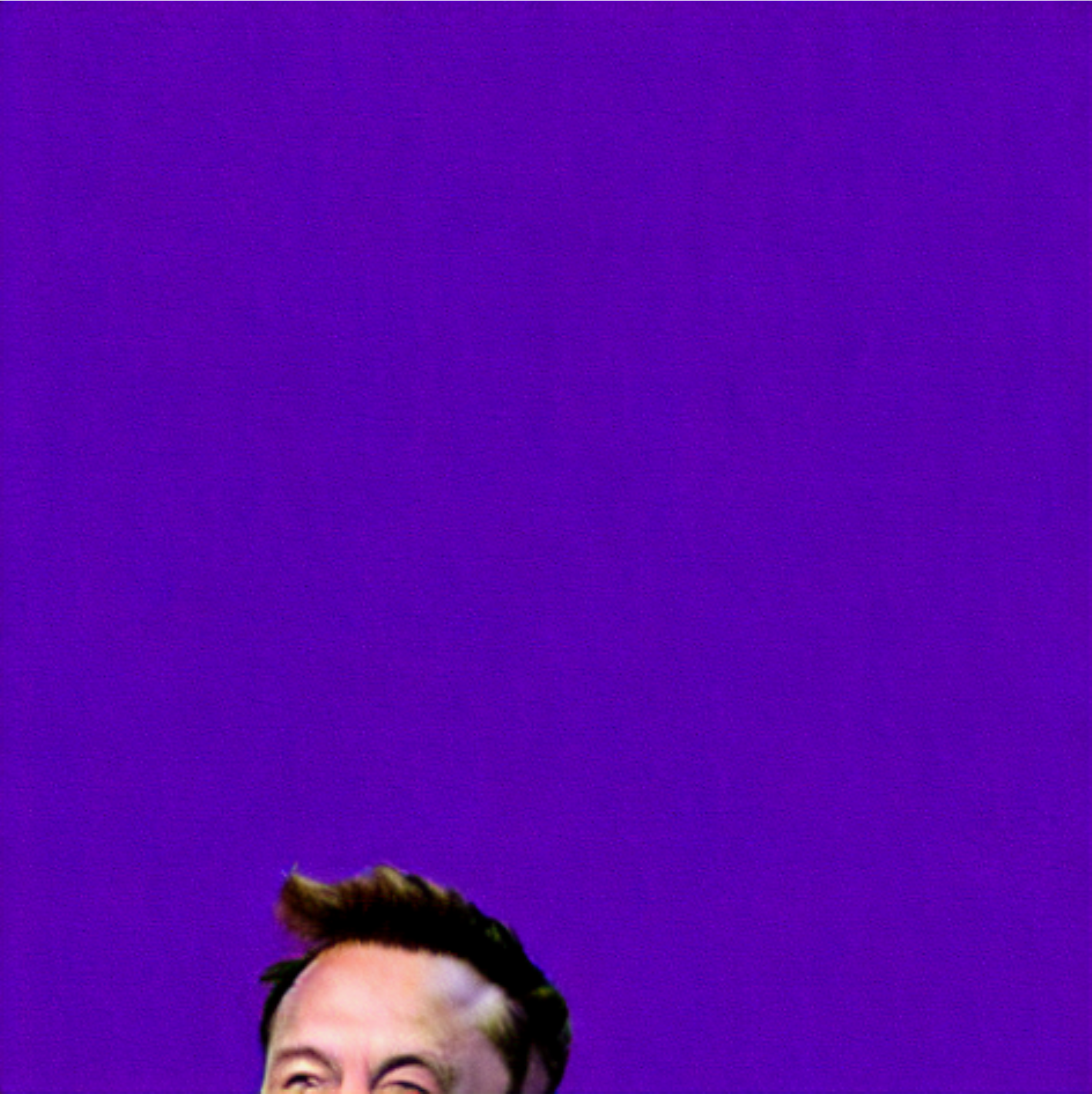} & 
\includegraphics[width=\mywidth]{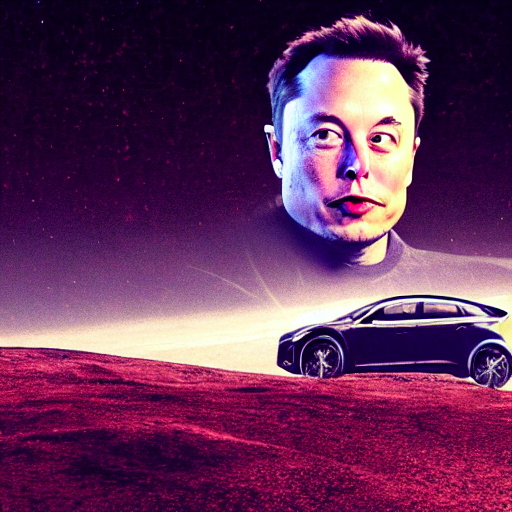} & 
\includegraphics[width=\mywidth]{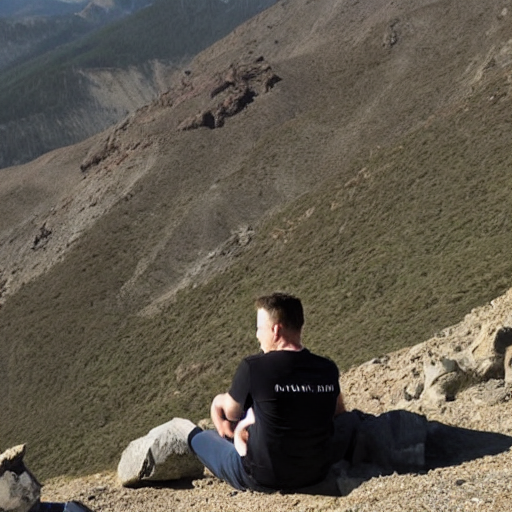} & 
\includegraphics[width=\mywidth]{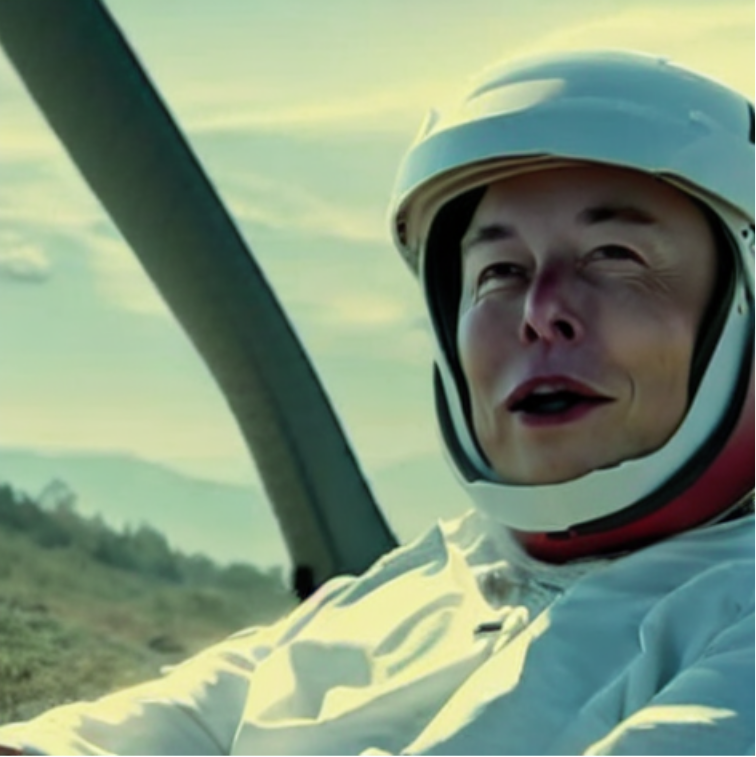} & 
\includegraphics[width=\mywidth]{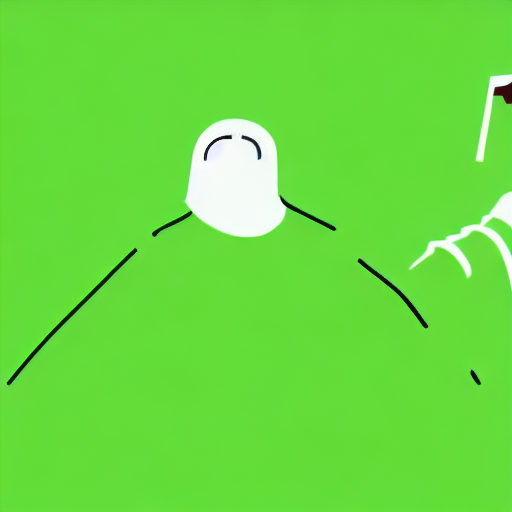} 
\\ 
\includegraphics[width=\mywidth]{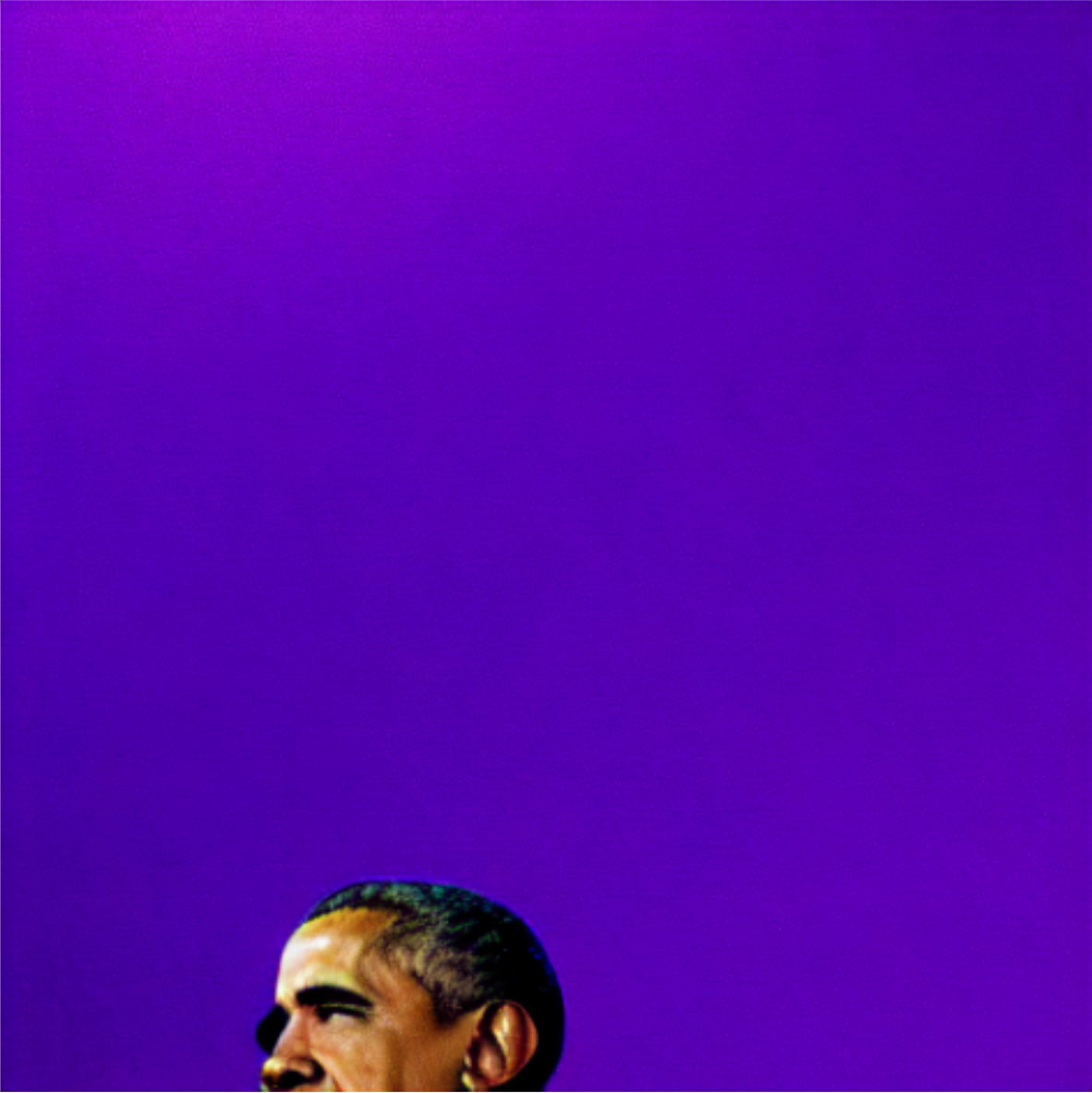} & 
\includegraphics[width=\mywidth]{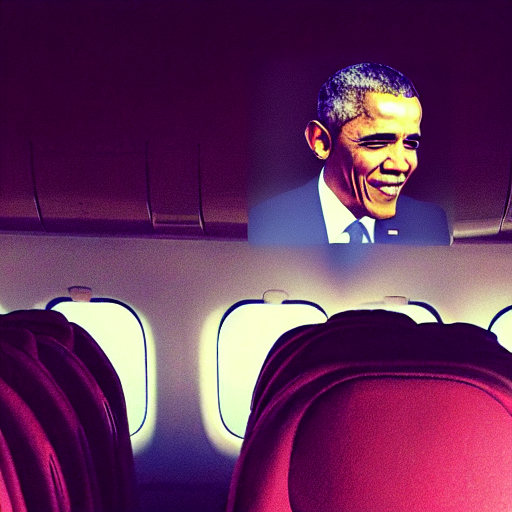} & 
\includegraphics[width=\mywidth]{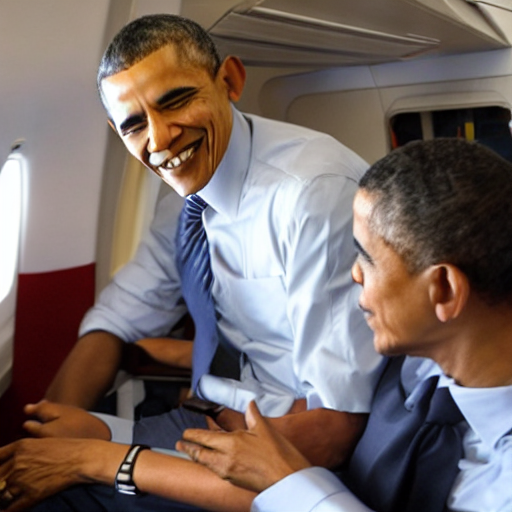} & 
\includegraphics[width=\mywidth]{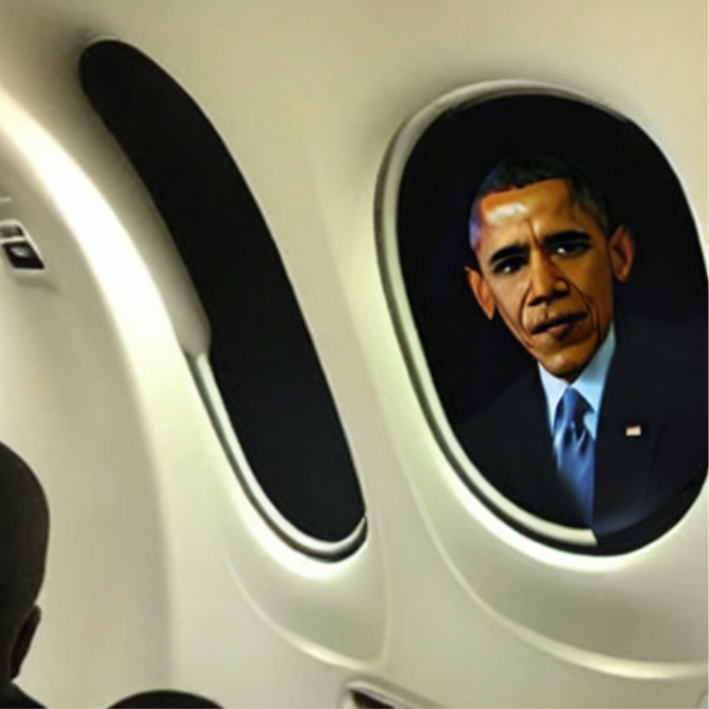} & 
\includegraphics[width=\mywidth]{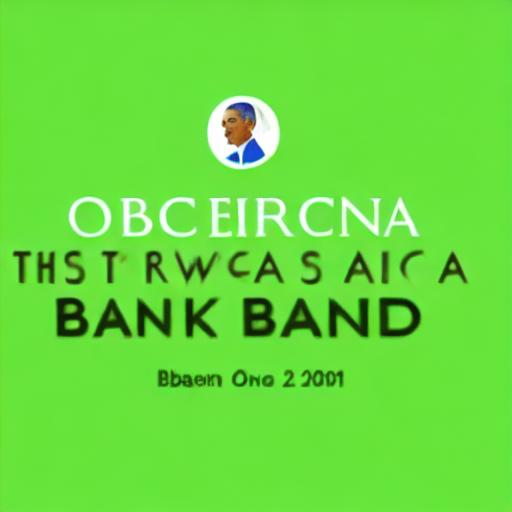} 
    \end{tabular}
    \caption{Samples generated with identical sampled noise and varying latent drift ($\delta \in [-0.1,0.1]$) at inference, with different text prompts using the pre-trained Stable Diffusion \cite{rombach2022high}. Top: \textit{Elon Musk on a mountain}, Bottom: \textit{Barack Obama on a plane}.}
    \label{fig:obama_musk}
\end{figure}

Although such models are trained on all types of data, their main objective is to produce photorealistic images, and the diversity is not limited by any means. This often contradicts  the nature of medical images, which are restricted to certain shape templates; for instance, bony structures like a skull are required to keep their shape, while soft tissues can be altered due to aging or disease progression. Fine-tuning techniques for diffusion models have been recently developed \cite{ruiz2023dreambooth,kumari2023multi,gal2022image}, in which a minute number of images (3-5) are used to introduce the new concept to the model; few works have investigated the usability of such methods for medical image generation \cite{de2023medical,chambon2022adapting,ruiz2023dreambooth} without success for generating brain Magnetic Resonance Images (MRIs).

In this work, we introduce a generalized formal definition of conditioning in diffusion models and pose it as a counterfactual explanation optimization problem \cite{wachter2017counterfactual}, which is a more restrictive way of conditioning suited for medical applications, see some examples in \cref{fig:interpret}. We demonstrate that by employing a pre-trained diffusion model at the inference time, without any fine-tuning, one can generate similarities by fixing the noise distribution as a conditioning factor from different text prompts, as can be seen in \cref{fig:obama_musk}. Referred to as \methodname{} (\method{}), our proposed method poses a min-max optimization problem, aiming to match the learned distribution of pre-trained models to a new distribution represented by finite accessible samples without accessing their training data when fine-tuning diffusion models. The latent space adds to the traditional conditions, e.g., text or image, and its underlying distribution functions as an additional hyperparameter and a conditioning factor. This would enable the model to make a trade-off between diversity and the desired condition while training. By employing \methodname{}, the model can adapt to the new domain more efficiently. \methodname{} boosts the image generation performance of SOTA approaches in medical image generation and manipulation using diffusion models by a large margin., and it closes the distribution gap between the pre-trained models dataset(s) and the target medical domain in contrast to the prior work in medical imaging with diffusion models \cite{khader2022medical, peng2022generating, pinaya2022brain, yoon2023sadm} and Generative Adversarial Networks (GANs) \cite{Jung.2021,pombo2023equitable,ravi2019degenerative,ravi2022degenerative,Xia.2021} which were trained from scratch.  %

The contributions of this work are as follows: (1) We propose a formal definition of conditioning for diffusion models based on the min-max optimization of a counterfactual formulation suitable for fine-tuning models on medical imaging data, (2) Our proposed \methodname{} is agnostic to the fine-tuning approaches and can be flexibly adapted for enhancing the distribution matching of pre-trained models on arbitrary data to the target data distribution, (3) We generate counterfactual medical images conditioned on text and image while preserving image quality and fidelity to the conditions. In addition, classifiers trained on synthetic images generated by \method{} show improved test accuracy on the real datasets, (4) We evaluate \methodname{} on two longitudinal brain MR datasets \cite{Weiner.2012,lamontagne2019oasis}, and a Chest-Xray dataset \cite{irvin2019chexpert}. These datasets contain different stages of patients, which makes them a fit for counterfactual image generation evaluations.

\section{Related Work}
\label{sec:literature}
\noindent\textbf{Image generation.}
Image generation is a task that involves creating realistic and diverse images either unconditionally  \cite{karras2020analyzing,karras2021alias} or from various inputs, such as text, sketches, or graphs \cite{farshad2023scenegenie}. It has many applications, such as image editing, synthesis, and manipulation \cite{dhamo2020semantic,farshad2022dispositionet,yeganeh2023anatomy,astaraki2023autopaint}. Image generation has been largely driven by GANs \cite{goodfellow2014generative} and diffusion models \cite{nichol2021improved}. In particular, conditional image generation \cite{mirza2014conditional} has been explored, which allows controlling the content and style of the generated images based on different modalities. Conditional image generation methods are generally categorized into the following: Image-to-image translation transforming an image from one domain to another, such as Pix2Pix \cite{isola2017image} and CycleGAN \cite{zhu2017unpaired}; Semantic image generation, which produces images from an input semantic map \cite{wang2018high,park2019SPADE}; Layout-based image generation generates images from bounding boxes and class labels for each scene instance \cite{zhao2018image, Sun_2019_ICCV}; and text-conditioned generative models \cite{
li2019storygan,nichol2022glide}. Recently, image generation has been revolutionized by the introduction of LDMs \cite{rombach2022high}, which enables unconditional and conditional high-resolution image generation and editing.

\vspace{0pt}
\noindent\textbf{Image Editing.}
Another relevant task is image manipulation \cite{dhamo2020semantic,jahoda2023prism}, which allows modifying an existing image according to user input (\textit{e.g.}, a mask, a sketch, or a graph). Image manipulation can be used for various purposes, such as removing or adding objects, changing colors or styles, or enhancing details. Image editing can be guided by text \cite{brooks2023instructpix2pix}, image \cite{parmar2023zero}, or semantics \cite{ntavelis2020sesame, ling2021editgan}.  InstructPix2Pix \cite{brooks2023instructpix2pix} is a recent approach that extends LDM \cite{rombach2022high} with an additional input channel to the U-Net for the latent representation of a source image.  Pix2Pix-Zero \cite{parmar2023zero} proposes a zero-shot text-conditioned image-to-image translation approach using LDM that does not rely on image pairs for training. 

\vspace{0pt}
\noindent\textbf{Medical Counterfactual Image Generation.}
Counterfactual image generation is a technique used to generate images that are similar to an existing image but have specific features altered to explore hypothetical scenarios \cite{sauer2020counterfactual,chang2021towards}. The goal is to create images that are realistic and consistent with the original image while also introducing the desired changes. In medical image generation, cGANs have been primarily used to generate counterfactual images. Pombo \etal \cite{pombo2023equitable} developed a cGAN architecture to generate brain MR images conditioned on age and gender.  Xia \etal \cite{Xia.2021} and Ravi \etal \cite{ravi2022degenerative} generated counterfactual trajectories of brain aging. cGANs have also been utilized to generate brain MR images with Alzheimer's disease from healthy brain MR scans \cite{Jung.2021}. Recently, diffusion models have gained more attention for medical image generation \cite{MullerFranzes.2023}.  Khader \etal \cite{khader2022medical} used the latent diffusion model (LDM) \cite{rombach2022high} to generate realistically looking 3D brain MRIs with a resolution of $64^3$. %
Peng \etal \cite{peng2022generating} were able to generate MRIs with a resolution of $128^3$ without employing significantly more training data by using a multi-step generation process. %
Pinaya \etal \cite{pinaya2022brain} trained a LDM with over 30K brain MRIs of the UK Biobank. The model can generate high-resolution MRIs conditioned on brain volume, ventricular volume, and age. Although image-to-image counterfactual brain MR modeling has not been explored to the best of our knowledge, there is limited work on predicting brain aging (\eg, \cite{yoon2023sadm}). They developed an architecture that predicts brain aging by conditioning a diffusion model on a longitudinal sequence of brain MRIs. 

\vspace{0pt}
\noindent\textbf{LDM Fine-Tuning.}
Most LDM-based models are trained on natural images. To help with the customization of the models, various fine-tuning techniques have been developed to allow a lightweight introduction of new concepts to the model \cite{choi2023custom}. %
Textual Inversion \cite{gal2022image} only fine-tunes the text encoder of the LDM architecture. The new concept to be learned is represented by a placeholder string $S_*$.  Instead of modifying the embedding space of the text encoder, Custom Diffusion \cite{gal2022image}, and DreamBooth \cite{ruiz2023dreambooth} focus on the tuning of the denoising U-Net. DreamBooth identifies rare tokens having a low prior in representing the new concept, and the "Class-specific Prior Preservation Loss" prevents the model from forgetting syntactic and semantic knowledge during fine-tuning. Custom Diffusion only fine-tunes the weights in the cross-attention layers in the denoising U-Net that act on the text conditioning. Limited work has been done on using those methods to introduce medical concepts to the pre-trained diffusion models \cite{yeganeh2024visage,lupke2024physics}. Chambon \etal \cite{chambon2022adapting} examined tuning different LDM elements for chest X-ray generation while generating prostate MR and chest X-ray images using Textual Inversion was explored in \cite{de2023medical}.

\def\mywidth{32pt} %

\begin{figure*}[tb]
\centering
 \begin{subfigure}[t]{0.48\textwidth}

\begin{tabular}{c@{\hspace{0.1em}}c@{\hspace{0.1em}}c@{\hspace{0.1em}}c@{\hspace{0.1em}}c@{\hspace{0.1em}}c@{\hspace{0.1em}}c@{\hspace{0.1em}}}
\multicolumn{1}{c}{-0.2} & \multicolumn{1}{c}{-0.1} & \multicolumn{1}{c}{-0.05} & \multicolumn{1}{c}{0} & \multicolumn{1}{c}{0.05} & \multicolumn{1}{c}{0.1} & \multicolumn{1}{c}{0.2} \\ 

\includegraphics[width=\mywidth]{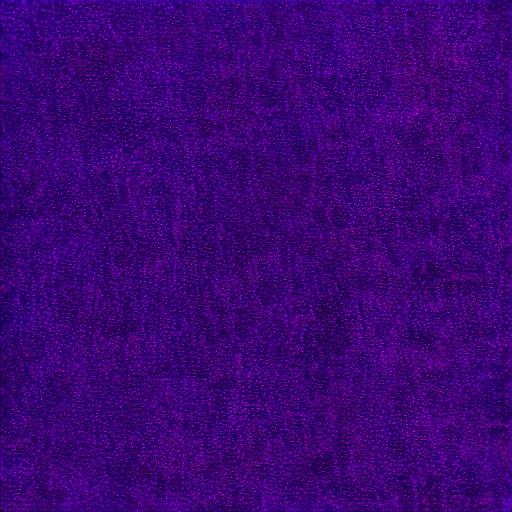} & \includegraphics[width=\mywidth]{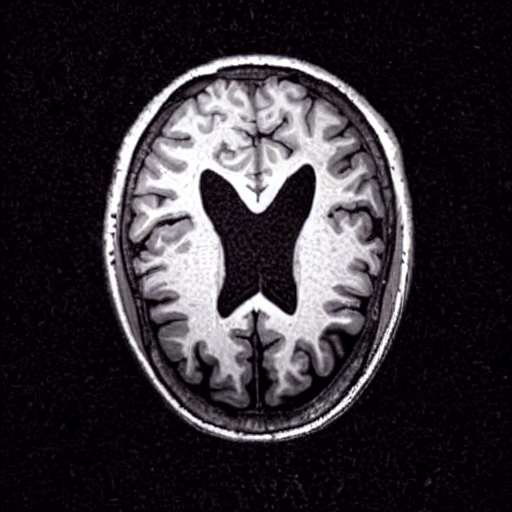} & \includegraphics[width=\mywidth]{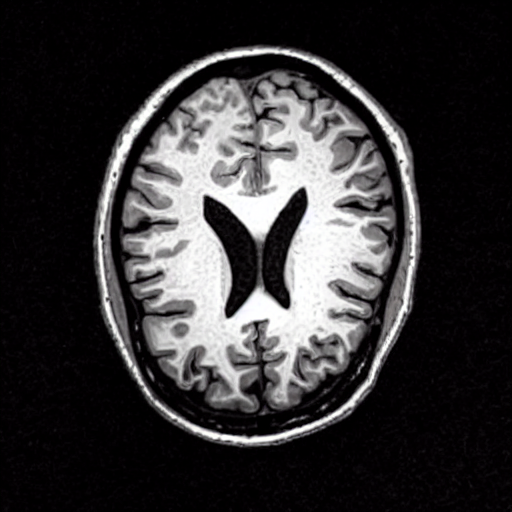} & \includegraphics[width=\mywidth]{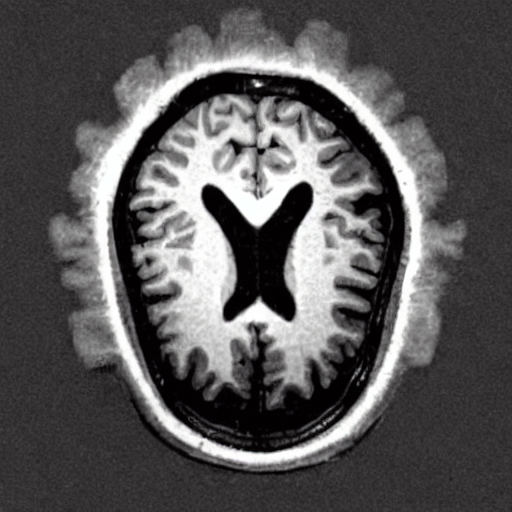} & \includegraphics[width=\mywidth]{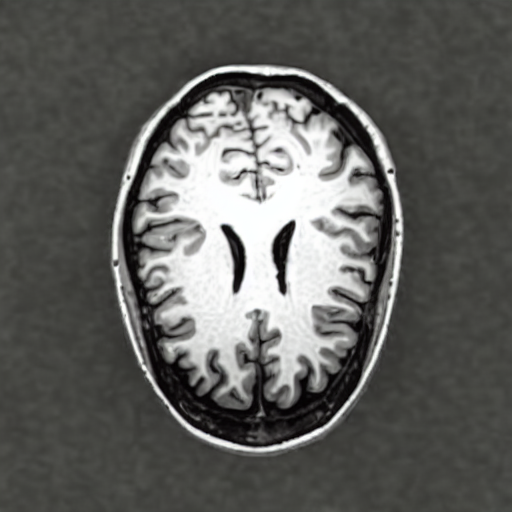} & \includegraphics[width=\mywidth]{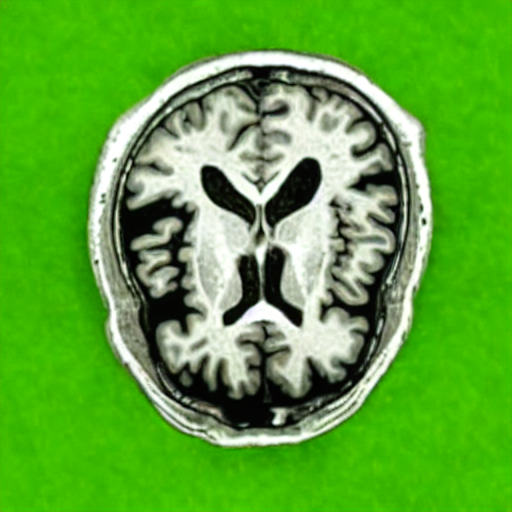} & \includegraphics[width=\mywidth]{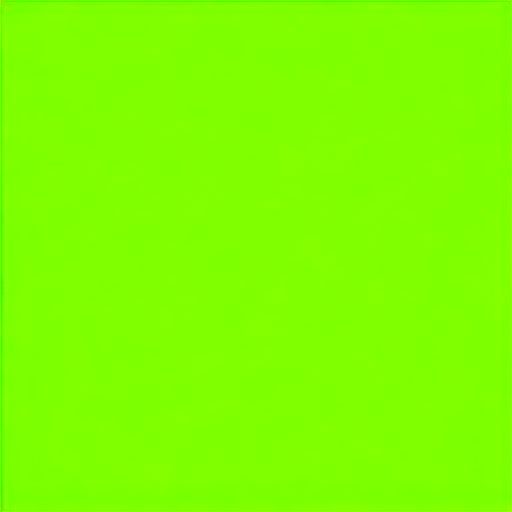}\\ 
 \includegraphics[width=\mywidth]{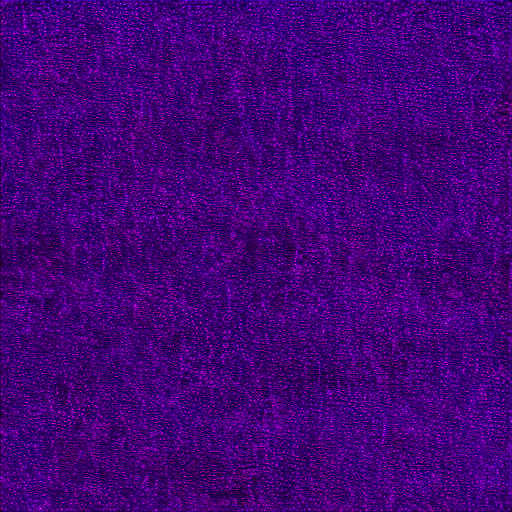} & \includegraphics[width=\mywidth]{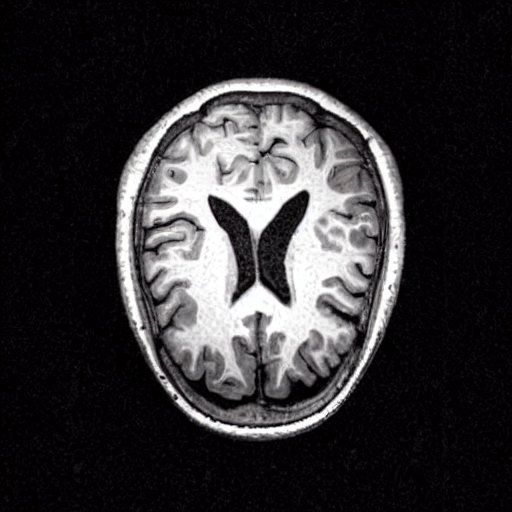} & \includegraphics[width=\mywidth]{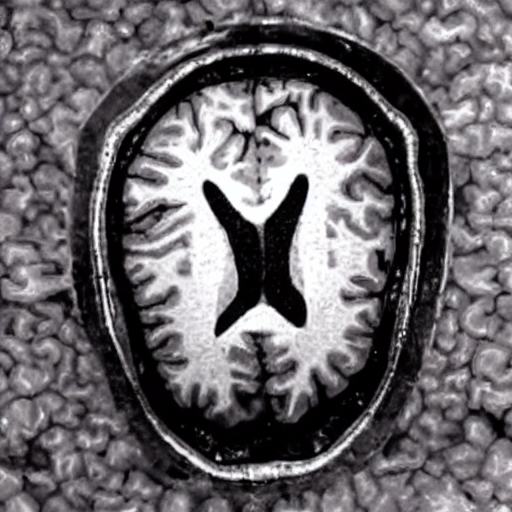} & \includegraphics[width=\mywidth]{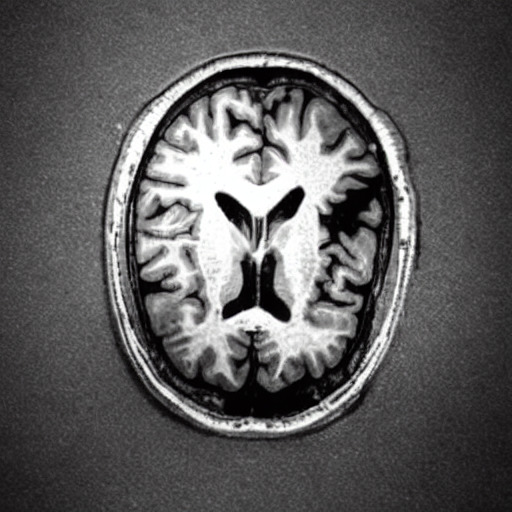} & \includegraphics[width=\mywidth]{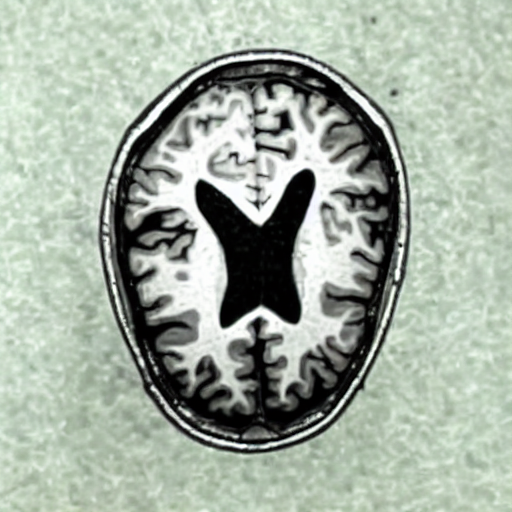} & \includegraphics[width=\mywidth]{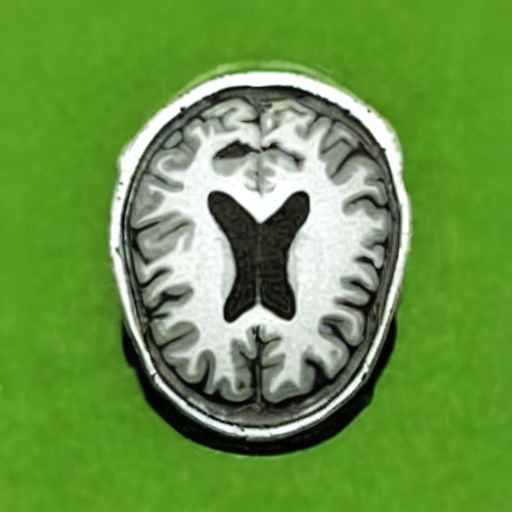} & \includegraphics[width=\mywidth]{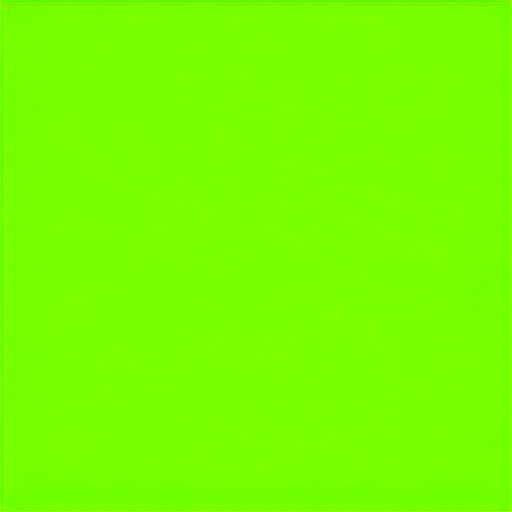} \\
      \includesvg[width=\mywidth]{images/example_images_tables/finetuned_CN_ld0_channel_wise_mean_offset-0.2.svg} & \includesvg[width=\mywidth]{images/example_images_tables/finetuned_CN_ld0_channel_wise_mean_offset-0.1.svg} & \includesvg[width=\mywidth]{images/example_images_tables/finetuned_CN_ld0_channel_wise_mean_offset-0.05.svg} & \includesvg[width=\mywidth]{images/example_images_tables/finetuned_CN_ld0_channel_wise_mean_offset0.svg} & \includesvg[width=\mywidth]{images/example_images_tables/finetuned_CN_ld0_channel_wise_mean_offset0.05.svg} & \includesvg[width=\mywidth]{images/example_images_tables/finetuned_CN_ld0_channel_wise_mean_offset0.1.svg} & \includesvg[width=\mywidth]{images/example_images_tables/finetuned_CN_ld0_channel_wise_mean_offset0.2.svg} \\ \includesvg[width=\mywidth]{images/example_images_tables/finetuned_CN_ld0_latent_means_offset_-0.2.svg} & \includesvg[width=\mywidth]{images/example_images_tables/finetuned_CN_ld0_latent_means_offset_-0.1.svg} & \includesvg[width=\mywidth]{images/example_images_tables/finetuned_CN_ld0_latent_means_offset_-0.05.svg} & \includesvg[width=\mywidth]{images/example_images_tables/finetuned_CN_ld0_latent_means_offset_0.svg} & \includesvg[width=\mywidth]{images/example_images_tables/finetuned_CN_ld0_latent_means_offset_0.05.svg} & \includesvg[width=\mywidth]{images/example_images_tables/finetuned_CN_ld0_latent_means_offset_0.1.svg} & \includesvg[width=\mywidth]{images/example_images_tables/finetuned_CN_ld0_latent_means_offset_0.2.svg} \\
\end{tabular}
\caption{SD \cite{rombach2022high} + Basic FT w/o \methodname{}}
 \end{subfigure} \hfill
 \begin{subfigure}[t]{0.48\textwidth}
 \begin{tabular}{c@{\hspace{0.1em}}c@{\hspace{0.1em}}c@{\hspace{0.1em}}c@{\hspace{0.1em}}c@{\hspace{0.1em}}c@{\hspace{0.1em}}c@{\hspace{0.1em}}}
\multicolumn{1}{c}{-0.2} & \multicolumn{1}{c}{-0.1} & \multicolumn{1}{c}{-0.05} & \multicolumn{1}{c}{0} & \multicolumn{1}{c}{0.05} & \multicolumn{1}{c}{0.1} & \multicolumn{1}{c}{0.2} \\ 
\includegraphics[width=\mywidth]{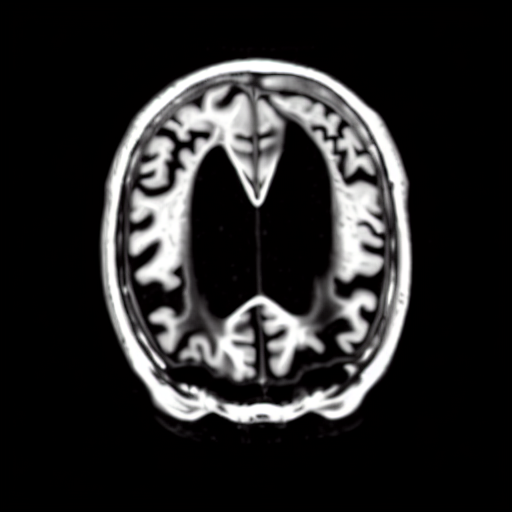} & \includegraphics[width=\mywidth]{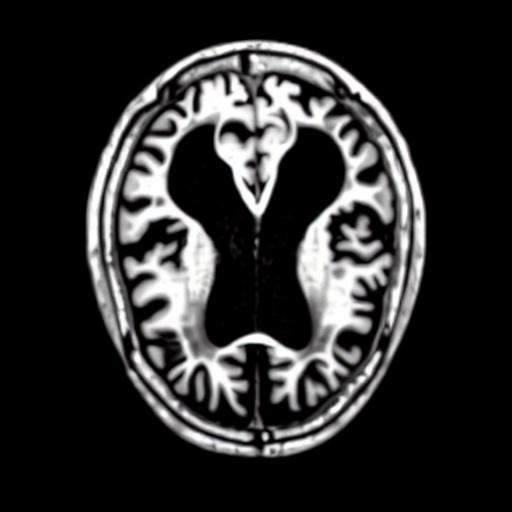} & \includegraphics[width=\mywidth]{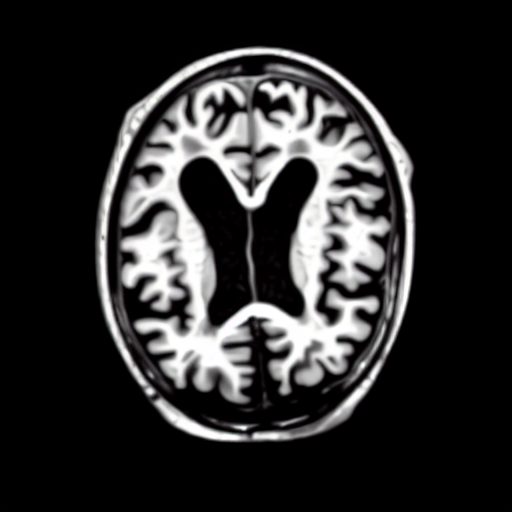} & \includegraphics[width=\mywidth]{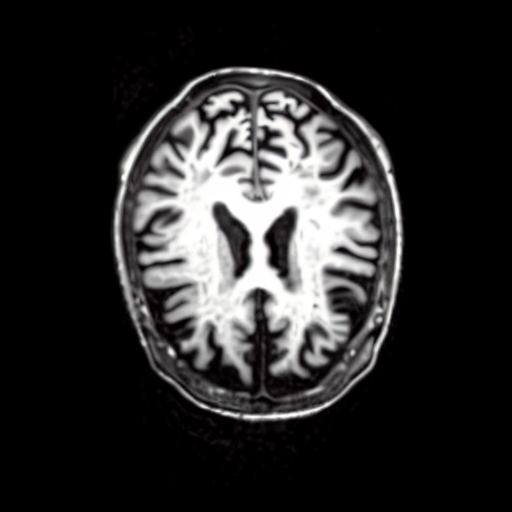} & \includegraphics[width=\mywidth]{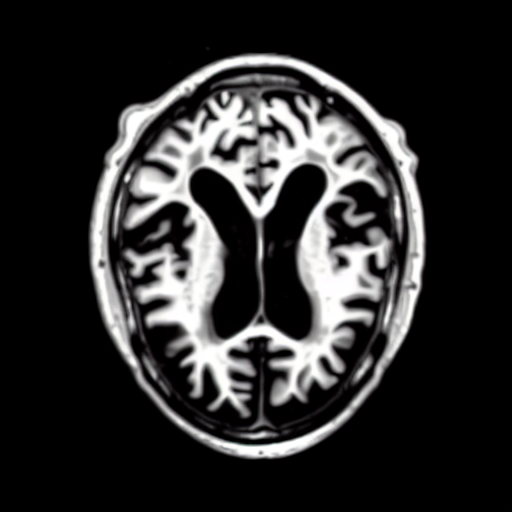} & \includegraphics[width=\mywidth]{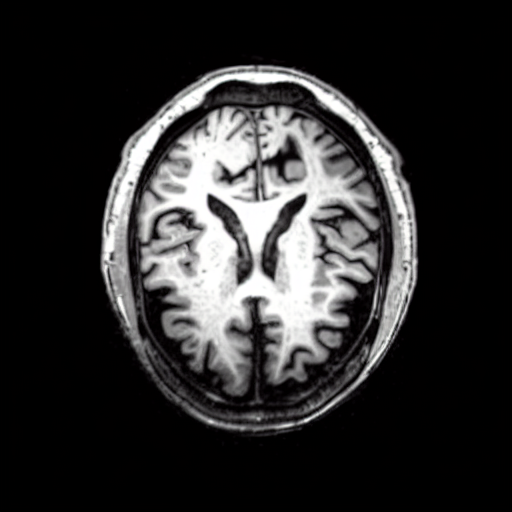} & \includegraphics[width=\mywidth]{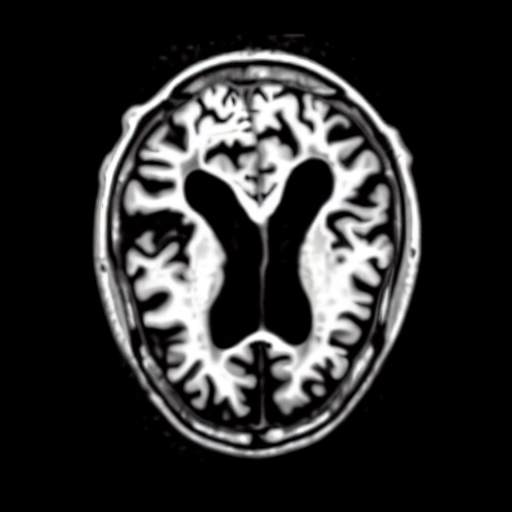} \\
\includegraphics[width=\mywidth]{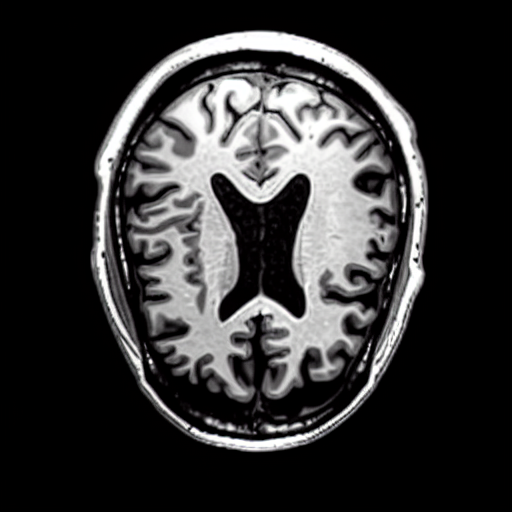} & \includegraphics[width=\mywidth]{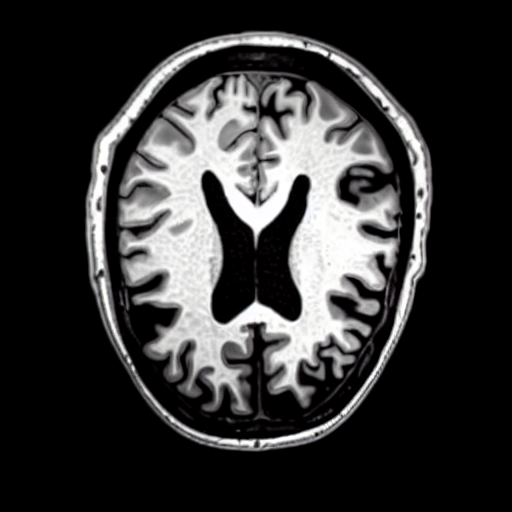} & \includegraphics[width=\mywidth]{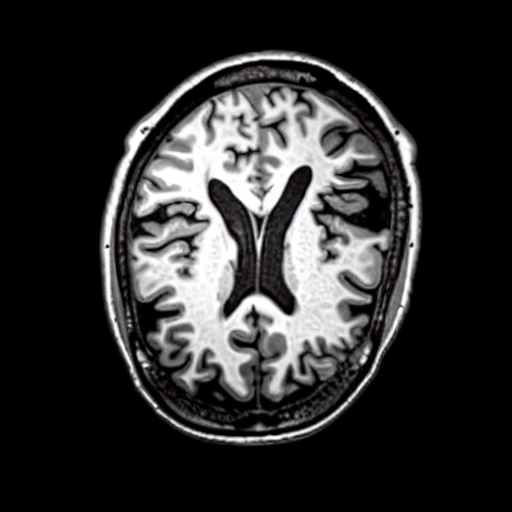} & \includegraphics[width=\mywidth]{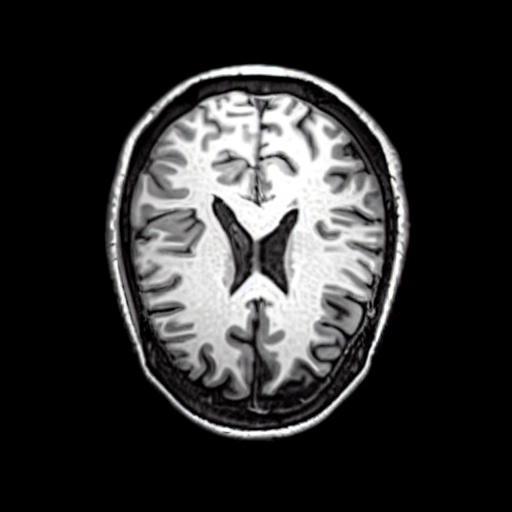} & \includegraphics[width=\mywidth]{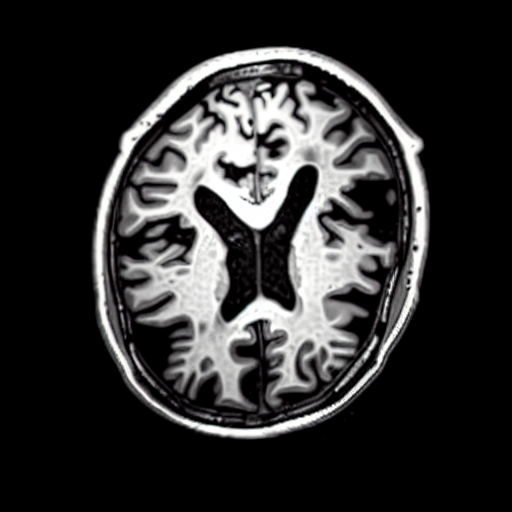} & \includegraphics[width=\mywidth]{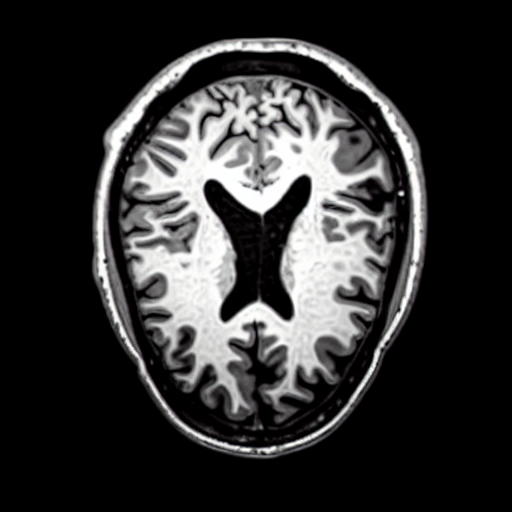} & \includegraphics[width=\mywidth]{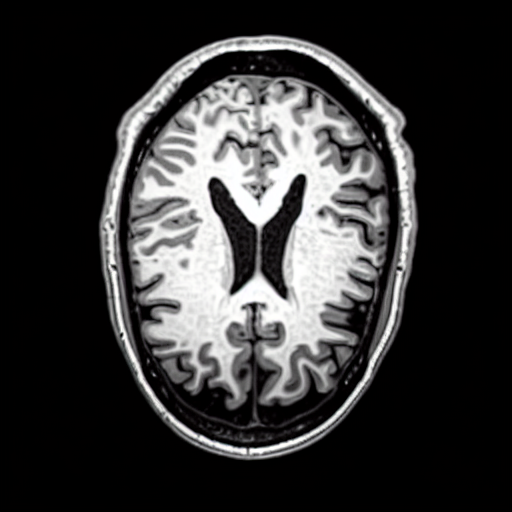} \\
\includesvg[width=\mywidth]{images/example_images_tables/finetuned_CN_channel_wise_mean_offset-0.2.svg} & \includesvg[width=\mywidth]{images/example_images_tables/finetuned_CN_channel_wise_mean_offset-0.1.svg} & \includesvg[width=\mywidth]{images/example_images_tables/finetuned_CN_channel_wise_mean_offset-0.05.svg} & \includesvg[width=\mywidth]{images/example_images_tables/finetuned_CN_channel_wise_mean_offset0.svg} & \includesvg[width=\mywidth]{images/example_images_tables/finetuned_CN_channel_wise_mean_offset0.05.svg} & \includesvg[width=\mywidth]{images/example_images_tables/finetuned_CN_channel_wise_mean_offset0.1.svg} & \includesvg[width=\mywidth]{images/example_images_tables/finetuned_CN_channel_wise_mean_offset0.2.svg} \\ %
\includesvg[width=\mywidth]{images/example_images_tables/finetuned_CN_latent_means_offset_-0.2.svg} & \includesvg[width=\mywidth]{images/example_images_tables/finetuned_CN_latent_means_offset_-0.1.svg} & \includesvg[width=\mywidth]{images/example_images_tables/finetuned_CN_latent_means_offset_-0.05.svg} & \includesvg[width=\mywidth]{images/example_images_tables/finetuned_CN_latent_means_offset_0.svg} & \includesvg[width=\mywidth]{images/example_images_tables/finetuned_CN_latent_means_offset_0.05.svg} & \includesvg[width=\mywidth]{images/example_images_tables/finetuned_CN_latent_means_offset_0.1.svg} & \includesvg[width=\mywidth]{images/example_images_tables/finetuned_CN_latent_means_offset_0.2.svg} \\ %
\end{tabular}
\caption{SD \cite{rombach2022high} + Basic FT w. \methodname{}}
  \end{subfigure}
  \caption{\textbf{Image and Latent Space Distribution w. and w/o. \method{} in Fine-tuning + Sampling}. Rows 1-2: sampled images with different latent drift parameters ($\delta$) during the inference. Row 3: channel-wise distribution change in images during the reverse sampling process. Row 4: distribution of the latent space $z_0$ during in reverse sampling.}
  \label{fig:dist_vis}
\end{figure*}

\section{Method}
\label{sec:methods}
We propose \methodname{} for generating counterfactual medical images using general pre-trained diffusion models. The diffusion model is either conditioned on a text prompt for text-to-image generation or an image and text for image-to-image translation.  
We are given a dataset $\mathcal{D}$ of images $I$ and labels $c$, where $I, c \in \mathcal{D}$, and the labels $c$ define the counterfactual element which corresponds to disease vs. healthy, or the patient's information such as age, gender, etc. Using a pre-trained diffusion model \textit{DM}, our goal is to fine-tune the \textit{DM} for prompt-based medical image generation and manipulation tasks. The diffusion model is parameterized by $\theta$.

\subsection{Diffusion Model}
Given a data distribution $x_0 \sim q(x_0)$, the forward Markov process generates a sequence of random variables $x_1, x_2, \dots, x_T$ with transition kernel $q(x_t | x_{t-1})$. Then, based on the chain rule and the Markov property, we can factorize the joint distribution of $x_1, x_2, \dots, x_T$ conditioned on $x_0$, denoted as $q(x_1, \dots, x_T | x_0)$, into:

\begin{equation}
q(x_1, \dots, x_T | x_0) = \prod_{t=1}^T q(x_t | x_{t-1}).
\end{equation}

In the forward process, noise is introduced into data until the distribution of latent space matches the Gaussian noise distribution $p(x_T)$ with the mean of $\mu$ and variance of $\sigma$, allowing to obtain $x_t$ in models like DDPM \cite{ho2020denoising}, where $x_{t} = \sqrt{\bar{\alpha}_{t}} x_{0} + \sqrt{1 - \bar{\alpha}_{t}} \epsilon_{t}$ $\text{with } \alpha_{t}:= 1 - \beta_{t} \text{ and } \bar{\alpha}_{t}=\prod_{s=0}^t \alpha_{s}{t+1}
\text{ and } \epsilon \sim \mathcal{N}(0, I)
$. In the reverse process, for generating new data samples, diffusion models start by sampling a noise vector from the prior distribution $p(x_T)$, then gradually removing noise by running a learnable Markov chain in the reverse time direction. Specifically, the reverse Markov chain is parameterized by a prior distribution $p(x_T)$ and a learnable transition kernel $p_\theta(x_{t-1} | x_t)$. The learnable transition kernel $p_\theta(x_{t-1} | x_t)$ takes the form of

\begin{equation}
p_\theta(x_{t-1} | x_t) = \mathcal{N}(x_{t-1}; \mu_\theta(x_t, t), \Sigma_\theta(x_t, t)),
\end{equation}
where $\theta$ denotes the model parameters, and the mean $\mu_\theta(x_t, t)$ and variance $\Sigma_\theta(x_t, t)$ are parameterized by deep neural networks. With this reverse Markov chain in hand, we can generate a data sample $x_0$ by first sampling a noise vector $x_T \sim p(x_T)$, then iteratively sampling from the learnable transition kernel $x_{t-1} \sim p_\theta(x_{t-1} | x_t)$ until $t=1$.

\subsection{Conditioning}

The diffusion model \textit{DM} is trained on a denoising objective:

\begin{equation}
    \mathbb{E}_{x,c,\boldsymbol{\epsilon},t}{w_t \|\hat{x}_\theta(\alpha_t x + \sigma_t \epsilon, c) - x \|^2_2},
\end{equation}
where $(x, c)$ are data-conditioning pairs, $t \sim \mathcal{U}([0, 1])$, $\epsilon \sim \mathcal{N}(0, I)$, and $\alpha_t, \sigma_t, w_t$ are functions of $t$ that influence sample quality. Intuitively, $\hat{x}_\theta$ is trained to denoise $z_t = \alpha_t x + \sigma_t \epsilon$ into $x$ using a squared error loss, weighted to emphasize certain values of $t$. 
Sampling such that the ancestral sampler~\cite{ho2020denoising} and DDIM~\cite{song2020denoising} start from pure noise $z_1 \sim \mathcal{N}(0, I)$ and iteratively generate points $z_{t_1}, \dotsc, z_{t_T}$, where $1 = t_1 > \cdots > t_T = 0$, modeled by  $p_\theta(x_{t-1} | x_t,c)$.

\subsection{\methodname{}}

We define \method{} as a hyperparameter in diffusion models to adapt the learned distribution of a pre-trained model $\mathcal{D}_\mathcal{\theta}$ to a new data distribution $\mathcal{D}_\mathcal{GT}$. \method{} is represented by a signed scalar value $\delta$ and is introduced to the diffusion process. For the fine-tuning through \methodname{}, \method{} is added to the target $z_T$ of the forward process, as well as the reverse processes:
\vspace{-2pt}
\begin{equation}
    p_\theta(x_{t-1} | x_t) = \mathcal{N}(x_{t-1}; \mu_\theta(x_t, t) + \delta, \Sigma_\theta(x_t, t)).
\end{equation}

 For example, in \cref{fig:obama_musk} \textit{LD} is applied to reverse process only at inference time showcasing how it can control the style of synthetic images, and \cref{fig:dist_vis} to both forward and reverse process for fine-tuning.
The discrepancy between the generated data distribution $\mathcal{D}_\mathcal{\theta}$ and the target data distribution $\mathcal{D}_\mathcal{GT}$ can be quantified by a distance function $d(\mathcal{D}_\mathcal{GT},\mathcal{D}_\mathcal{\theta})$. In the absence of the training set used to fine-tune the pre-trained model, this distance can be estimated via Monte Carlo sampling of generated samples from $\mathcal{D}_\mathcal{\theta}$ and existing samples in $\mathcal{D}_\mathcal{GT}$. Here, we use $L1_{norm}$ as the distance function between the synthetic samples and the target dataset, as further explained in the algorithm in the supplement. %
The diversity of generated samples in a diffusion model is ensured by the stochastic element introduced by $\mathcal{N}(\mu_, \sigma)$. %
However, in the reverse Markov chain process of a conditional model $p_{\theta}(z_{t-1} | z_t, c)$, $\mathcal{N}(\mu_, \sigma)$ remains unchanged despite the introduction of condition c. This phenomenon is demonstrated in \cref{fig:dist_vis} by visualizing the generated samples, the channel-wise data distribution, and the latent space distribution with and without \methodname{}. As it can be seen, the data and latent distribution have a high variance when fine-tuned without \method{}, while the model with \method{} reaches a stable point that is resilient and robust to distribution shift. We hypothesize that the final latent variable  $z_T$ must be considered as part of the condition, and with our proposed $\delta$, we can modify $\mu$ to ensure that the learned representations of $\mathcal{D}_\mathcal{\theta}$ accurately reflect the target data distribution $\mathcal{D}_\mathcal{GT}$. 

\subsection{Counterfactual Image Generation}
A sample $x$ is labeled as $y=\hat{f}(x)$, such that $(x,y) \in \mathcal{D}_\mathcal{GT}$, and a counterfactual sample $x'$ is a synthetically generated by a generative model $G_{\theta}(x)$ that is highly similar to $x$ but has different enough features such that it can be labeled differently $y'=\hat{f}(x')$. This can be achieved by minimizing two contradictory loss values, resembling a min-max problem with the following objective function \cite{wachter2017counterfactual}:

\begin{align}\label{eq:counterf}
L(x, x', y', \lambda) = {} & \underset{\ell_{o}}{\text{min}} \underbrace{\left[ \lambda \cdot(\ell_{o}(\hat{f}(x'), y'))\right]}_{\text{Desired Outcome Fidelity}} \nonumber \\
& + \underset{\ell_{in}}{\text{min}}\underbrace{\left[ \ell_{in}(x, x')\right]}_{\text{Counterfactual Fidelity}}
\end{align}

where $\ell_{in}(x, x')$ ensures the similarity of the original instance $x$ and the counterfactual instance $x'$, and $\ell_{o}(\hat{f}(x') , y')$ is a term to minimize the model’s prediction for the counterfactual instance $\hat{f}(x')$ and the desired outcome $y'$. Inherently, $\ell_{in} \propto \frac{1}{\ell_{o}}$ based on the definition of counterfactual conditions. The weighting factor $\lambda$ controls the trade-off between achieving the desired outcome and maintaining similarity to the original instance; hence, the model is optimized by minimizing $\ell_{in}$, which leads to $x \sim x'$. Conversely, higher values of $\lambda$, ($1$ in our counterfactual experiments) facilitate additional conditioning by introducing the latent drifting parameter $\delta$ affecting new sampled latent value $x'$ for the new data points.

\vspace{5pt}
\noindent \textbf{\methodname{} in Diffusion Models.}
A diffusion model with parameters \(\theta\), pre-trained on a large dataset of natural images $\mathcal{D}_\mathcal{\theta}$, generates samples by sampling from a latent space (\(z_T \sim \mathcal{N}(0, I)\)). Standard fine-tuning on a limited dataset of medical images $\mathcal{D}_\mathcal{GT}$ assumes that model parameters $\theta$ can directly transform to $\theta'$. However, the significant distribution shift between natural and medical images prevents the direct adaptation of generic foundation diffusion models. LD reframes this as a counterfactual generation problem, positing that $\theta$ and $\theta'$ belong to a more general domain. Therefore, in a large pre-trained model $\mathcal{D}_\mathcal{\theta}$ resembling a general domain, a latent space $z'$, similar to $z$, exists, such that the model generates samples from $\mathcal{D}_\mathcal{GT}$ when conditioned on $z'$.
In \cref{eq:counterf}, if (\(\lambda = 0\)), the \textit{Desired Outcome Fidelity} term is eliminated, resulting in (\(z = z'\)) (standard fine-tuning). However, if (\(\lambda > 0\)), the minimum value of \(\delta\) is found via grid search to minimize the distance function. Fine-tuning then maximizes the probability that the model produces results belonging to the target domain. Here, we employed the \(L1_{\text{norm}}\) as the distance function and tune hyperparameter $\delta$. %

\def\mywidth{50pt} %
\begin{figure*}[tb]
    \centering
    \begin{minipage}[c]{0.02\linewidth}
        \raggedright
        \rotatebox[origin=c]{90}{\quad CN \quad\quad\quad\quad\quad\quad\quad AD}
    \end{minipage}
    \begin{minipage}[c]{0.23\linewidth}
        \begin{subfigure}{\linewidth}
            \centering
            \includegraphics[width=\mywidth]{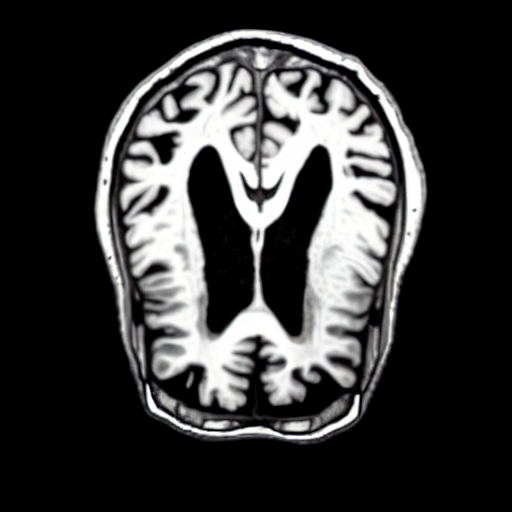}
            \includegraphics[width=\mywidth]{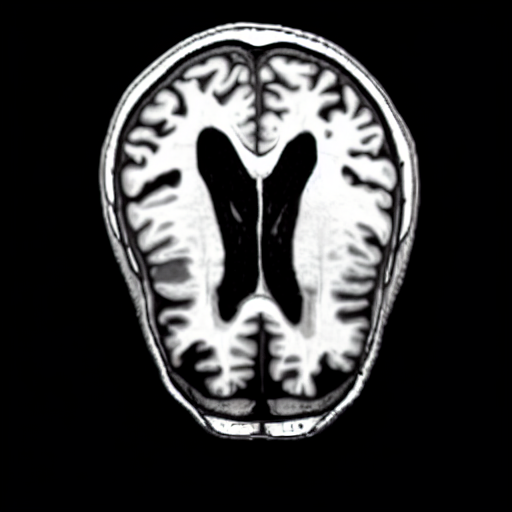}
            \vspace{2mm}
            \includegraphics[width=\mywidth]{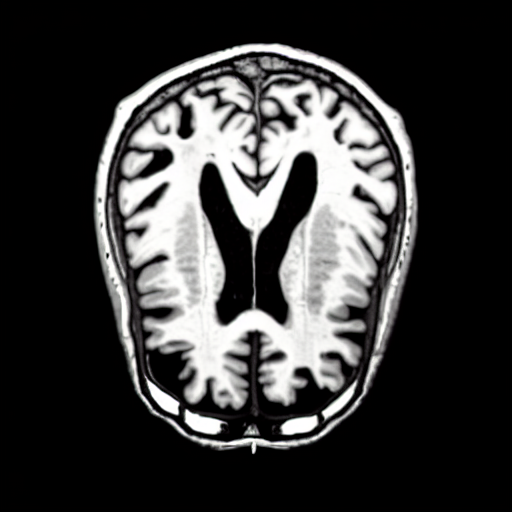}
            \includegraphics[width=\mywidth]{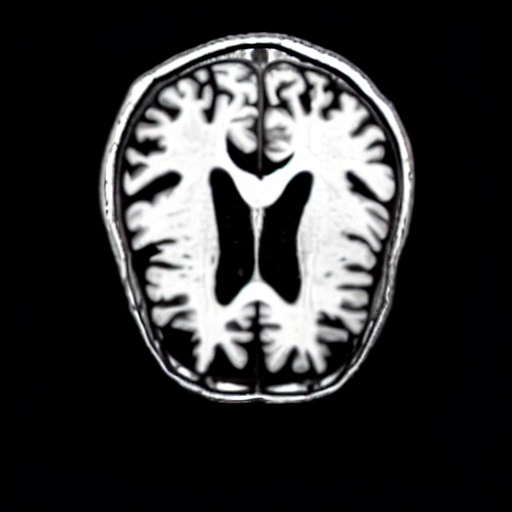}
            \includegraphics[width=\mywidth]{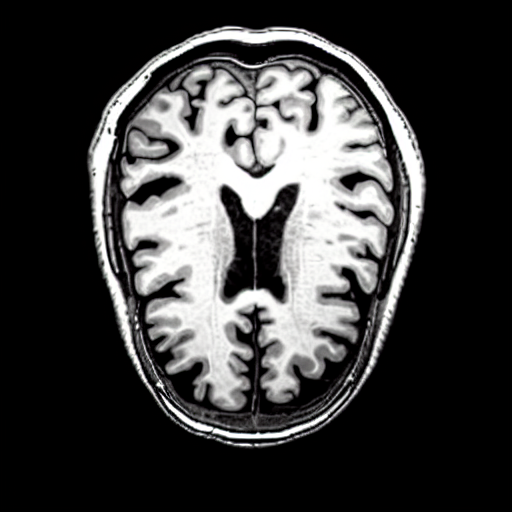}
            \includegraphics[width=\mywidth]{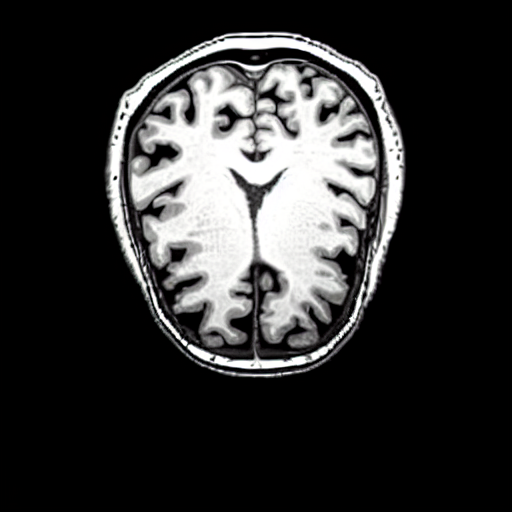}
            \includegraphics[width=\mywidth]{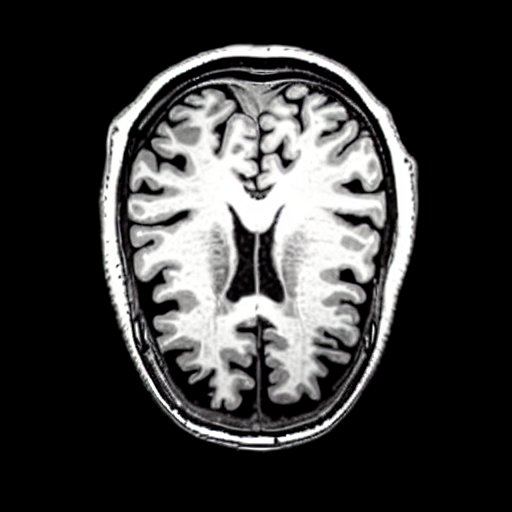}
            \includegraphics[width=\mywidth]{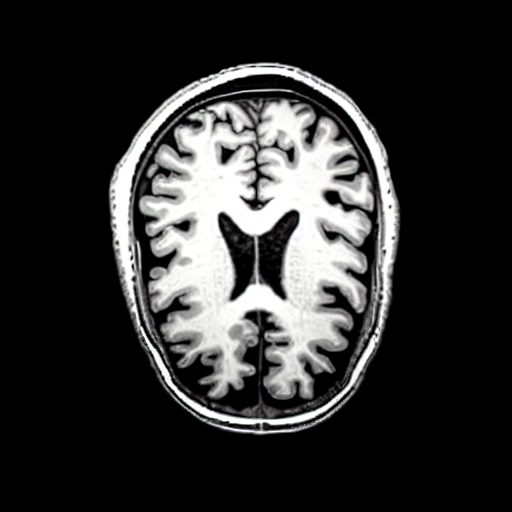}
            \caption{\tiny Textual Inversion \cite{gal2022image}}
        \end{subfigure}
    \end{minipage}
    \begin{minipage}[c]{0.23\linewidth}
        \begin{subfigure}{\linewidth}
            \centering
            \includegraphics[width=\mywidth]{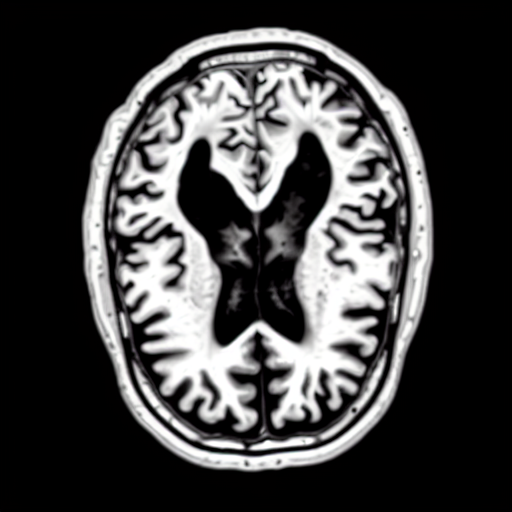}
            \includegraphics[width=\mywidth]{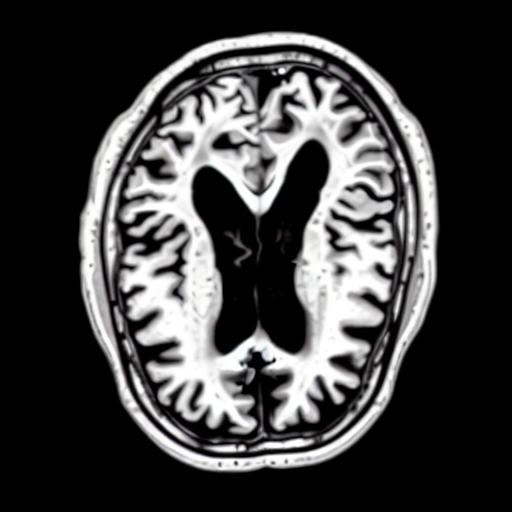}
            \vspace{2mm}
            \includegraphics[width=\mywidth]{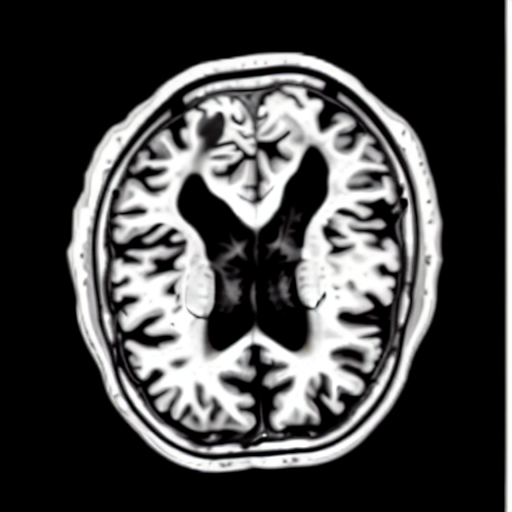}
            \includegraphics[width=\mywidth]{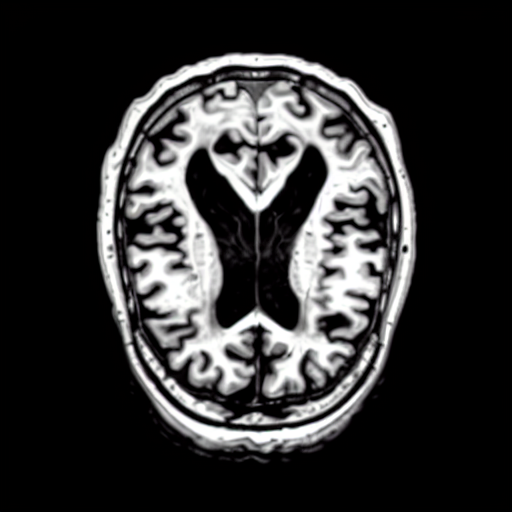}
            \includegraphics[width=\mywidth]{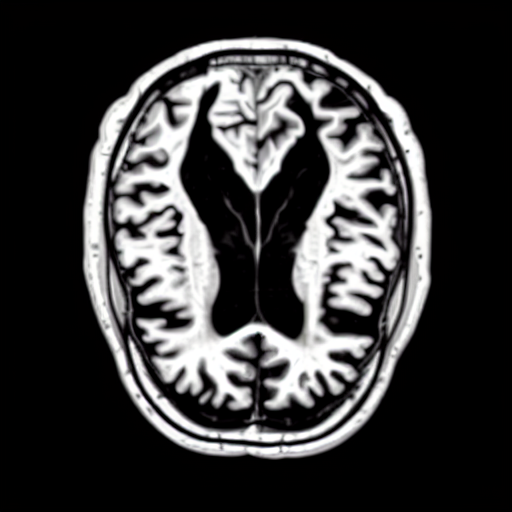}
            \includegraphics[width=\mywidth]{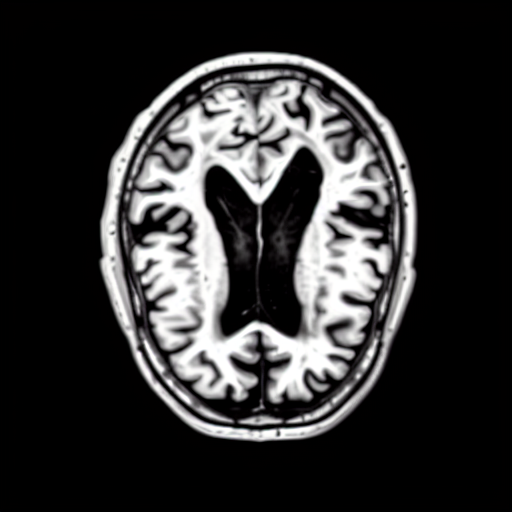}
            \includegraphics[width=\mywidth]{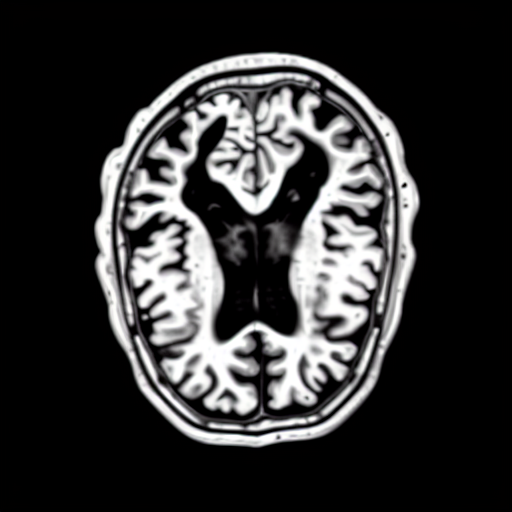}
            \includegraphics[width=\mywidth]{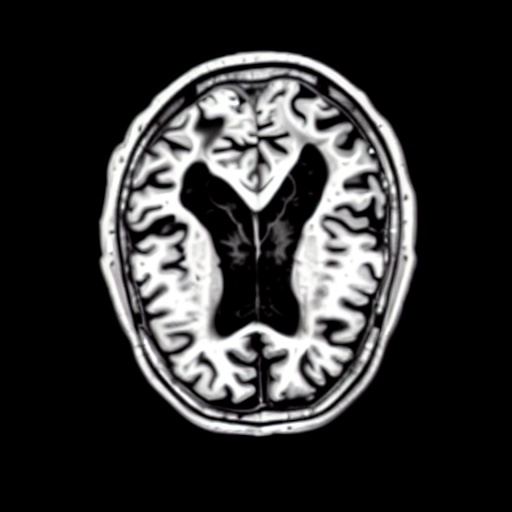}
            \caption{\tiny DreamBooth \cite{ruiz2023dreambooth}}
        \end{subfigure}
    \end{minipage}
    \begin{minipage}[c]{0.23\linewidth}
        \begin{subfigure}{\linewidth}
            \centering
            \includegraphics[width=\mywidth]{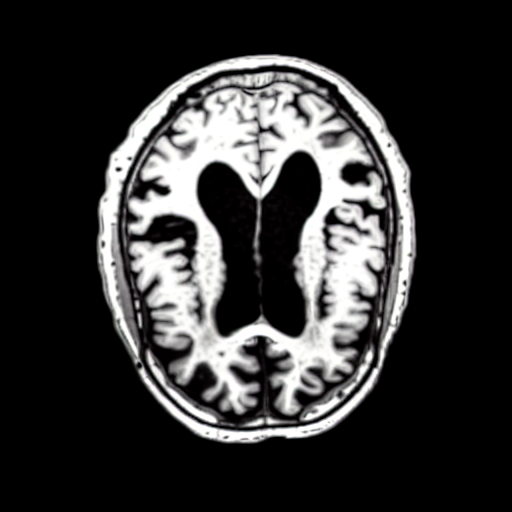}
            \includegraphics[width=\mywidth]{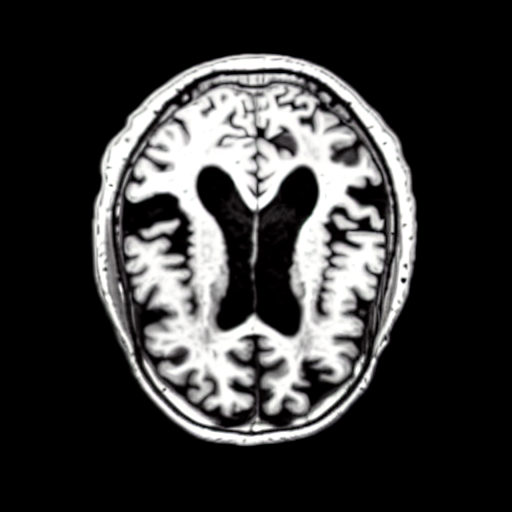}
            \vspace{2mm}
            \includegraphics[width=\mywidth]{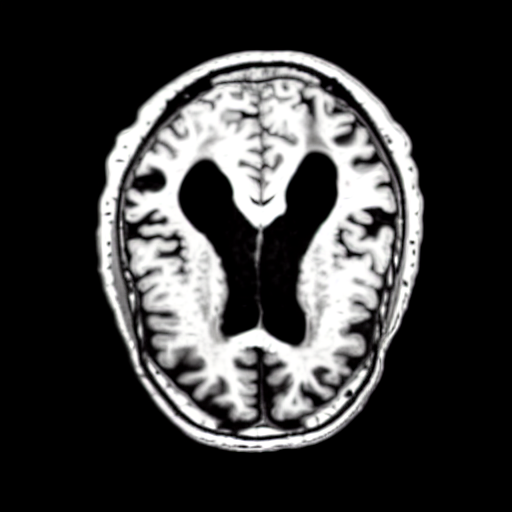}
            \includegraphics[width=\mywidth]{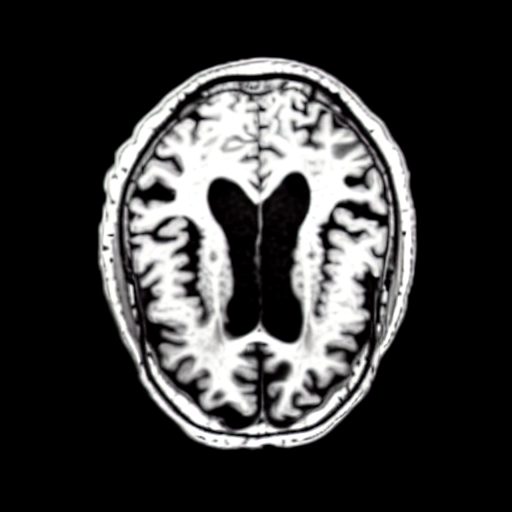}
            \includegraphics[width=\mywidth]{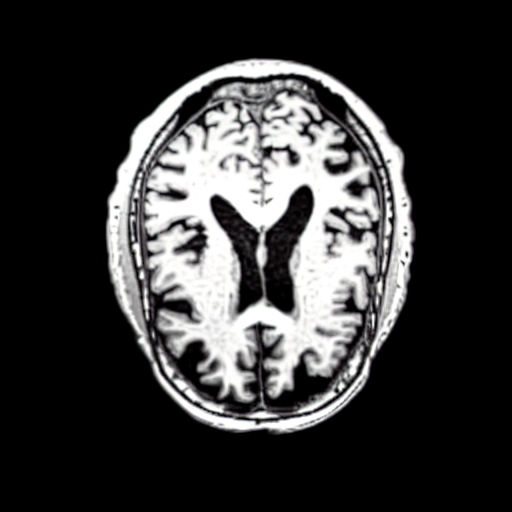}
            \includegraphics[width=\mywidth]{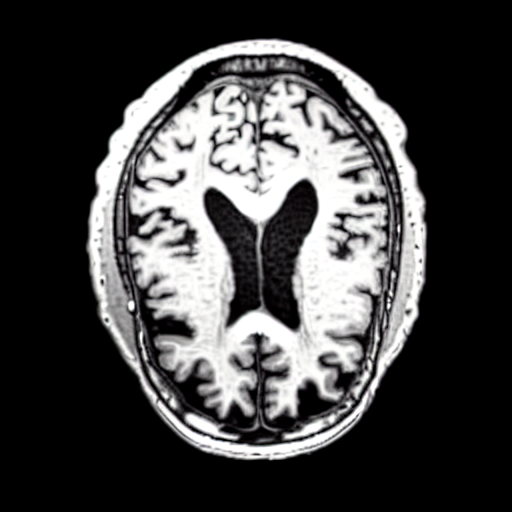}
            \includegraphics[width=\mywidth]{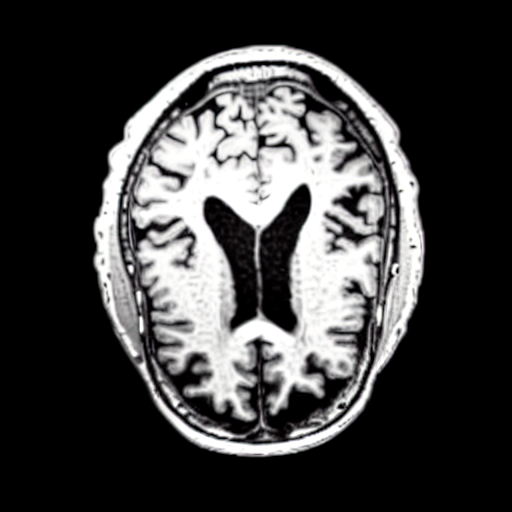}
            \includegraphics[width=\mywidth]{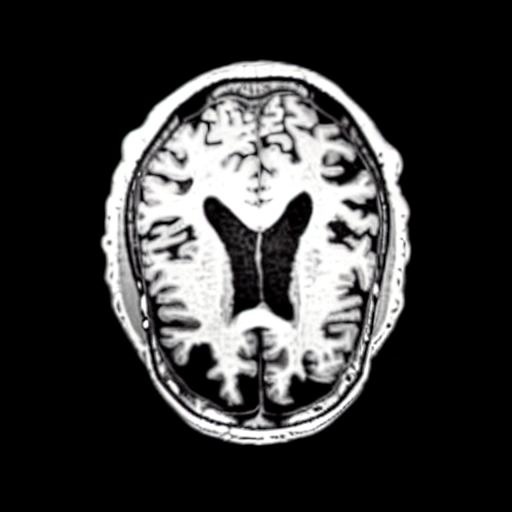}
            \caption{\tiny Custom Diffusion \cite{kumari2023multi}}
        \end{subfigure}
    \end{minipage}
    \begin{minipage}[c]{0.23\linewidth}
        \begin{subfigure}{\linewidth}
            \centering
            \includegraphics[width=\mywidth]{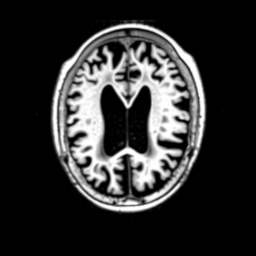}
            \includegraphics[width=\mywidth]{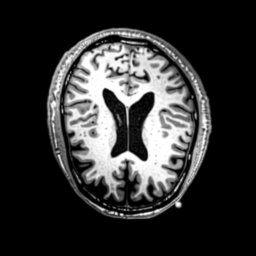}
            \vspace{2mm}
            \includegraphics[width=\mywidth]{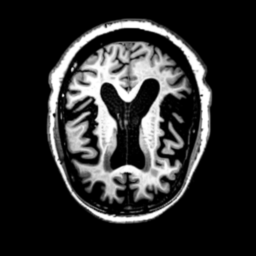}
            \includegraphics[width=\mywidth]{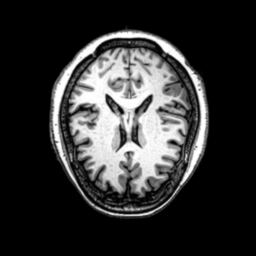}
            \includegraphics[width=\mywidth]{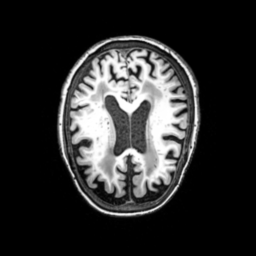}
            \includegraphics[width=\mywidth]{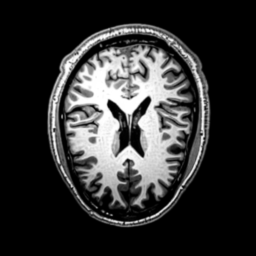}
            \includegraphics[width=\mywidth]{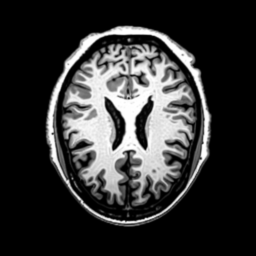}
            \includegraphics[width=\mywidth]{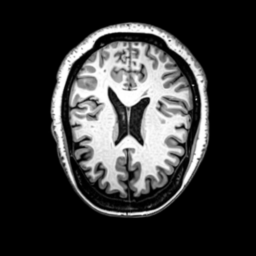}
            \caption{\tiny Basic FT \cite{rombach2022high}}
        \end{subfigure}
    \end{minipage}
    \caption{MRI Slice Generation for Cognitively Normal (CN) and Alzheimer's Disease (AD) after fine-tuning Stable Diffusion with \textbf{\method{}} using different methods.}
    \label{fig:finetuning_examples}
\end{figure*}

\begin{table*}[tb]
    \centering
    \caption{Comparison of different fine-tuning methods for text-to-image generation, with and without \textbf{\method{}}, on brain MRI \cite{Weiner.2012,lamontagne2019oasis} and chest X-ray \cite{irvin2019chexpert} data. Values from real data are provided as references. The number of samples is consistent across experiments, with Real+Synthetic using a $50\%$ split between real data and samples generated by Stable Diffusion \cite{rombach2022high} Basic FT with \method{}.}
    \begin{tabular}{l@{\hskip 0.4in} c@{\hskip 0.4in} c c c@{\hskip 0.4in}  c c c}
    \hline
    \multirow{2}{*}{Method} & \multirow{2}{*}{LD} & \multicolumn{3}{c}{Brain MR} & \multicolumn{3}{c}{CheXpert} \\ \cline{3-8}
    & & FID $\bm{\downarrow}$ & KID $\bm{\downarrow}$ & AUC $\bm{\uparrow}$ & FID $\bm{\downarrow}$ & KID $\bm{\downarrow}$ & AUC $\bm{\uparrow}$ \\
    \hline
    \multirow{2}{*}{Custom Diffusion \cite{kumari2023multi}} & \xmark & 129.21 & 0.132 & \underline{0.609} & 323 & 0.275 & 0.573 \\
    & \cmark & \underline{63.58} & \underline{0.065} & 0.544 & \underline{315} & \underline{0.270} & \underline{0.593} \\
    \hline
    \multirow{2}{*}{DreamBooth \cite{ruiz2023dreambooth}} & \xmark & 130.92 & 0.125 & 0.500 & 188 & 0.175 & 0.567 \\
    & \cmark & \underline{92.37} & \underline{0.099} & \underline{0.512} & \underline{177} & \underline{0.145} & \underline{0.582} \\
    \hline
    \multirow{2}{*}{Textual Inversion \cite{gal2022image}} & \xmark & 120.63 & 0.098 & 0.600 & 171.77 & 0.135 & 0.600 \\
    & \cmark & \underline{67.56} & \underline{0.065} & \underline{0.670} & \underline{133.18} & \underline{0.085} & \underline{0.640} \\
    \hline
    Stable Diffusion \cite{rombach2022high} & \xmark & 313.61 & 0.289 & 0.414 & 331 & 0.286 & 0.426 \\
    \multirow{2}{*}{\quad + Basic FT} & \xmark & 92.13 & 0.071 & 0.704 & 112 & 0.097 & 0.672 \\
    & \cmark & \underline{\textbf{49.68}} & \underline{\textbf{0.035}} & \underline{\textbf{0.724}} & \underline{\textbf{84}} & \underline{\textbf{0.077}} & \underline{\textbf{0.746}} \\
    \hline
    Real Data & - & - & - & 0.870 & - & - & 0.880 \\
    Real+Synthetic Data & \cmark & - & - & \textbf{0.883} & - & - & \textbf{0.892} \\
    \hline
    \end{tabular}
    \label{tab:combined_metrics}
\end{table*}

\section{Experiments}
\label{sec:experiments}
To show the effectiveness of our method in medical image generation and manipulation, we adopted several methods: Stable Diffusion \cite{rombach2022high} with various fine-tuning schemes, namely Custom Diffusion \cite{kumari2023multi}, Dreambooth \cite{ruiz2023dreambooth}, and Textual Inversion \cite{de2023medical} for text-to-image generation, and Pix2Pix Zero \cite{parmar2023zero}, and InstructPix2Pix \cite{brooks2023instructpix2pix} for text-conditioned image-to-image generation. 
\subsection{Experimental Setup}
For all the experiments, the SD-v1.4 model pre-trained on the LAION-5B dataset \cite{schuhmann2022laion} was employed.
The Stable Diffusion framework is based on three main elements: A text encoder (here based on CLIP \cite{radford2021learning}), a latent diffusion model consisting of an Autoencoder, and a text-conditioned U-Net. Training all of these elements can be computationally expensive. Thus, fine-tuning methods have been developed that focus on a particular element of the Stable Diffusion framework. We evaluate four fine-tuning methods with and without \method{}, namely textual inversion, DreamBooth, Custom Diffusion, and Stable Diffusion basic fine-tuning (fine-tuning the denoising U-Net while freezing the rest of the components). While the first three methods only require a couple of samples for introducing a new concept, the latter requires a large amount of data. The implementation details and dataset preprocessing are provided in the supplement. 

\vspace{5pt}
\noindent \textbf{Datasets.}
To fine-tune and evaluate our method, we utilize three medical imaging datasets. For the experiments on brain MR imaging, we
utilize the ADNI-1 \cite{Weiner.2012} and OASIS-3 \cite{lamontagne2019oasis} datasets which include longitudinal MR scans of Mild Cognitive Impairment (MCI), Alzheimer's Disease (AD) or Cognitively Normal (CN) patients. For the chest x-ray experiments, we used the CheXpert \cite{irvin2019chexpert} dataset, which is a large dataset containing over 224K chest radiographs of four categories: no finding (healthy), Cardiomegaly, Pleural Effusion, and Pneumonia. The datasets were preprocessed, and the final datasets consisted of $3269$ scans (414 AD, 634 MCI, 2214 CN) for the brain dataset and 800 uniformly random samples from the chest x-ray dataset. The details of data preprocessing are reported in the supplementary material. All models for brain MR generation are evaluated on 200 samples, and the models on chest X-rays are evaluated on 400 test samples.

\vspace{5pt}
\noindent \textbf{Evaluation Metrics.} For the evaluation of the image realism, we calculate the Fr\'echet Inception Distance (FID) \cite{heusel2017gans} and Kernel Inception Distance (KID) \cite{binkowski2018demystifying} between the synthetically generated samples and our test set. Additionally, we train binary classification models (CN/AD) on a Resnet18 architecture using 600 synthetically generated brain MR (300 AD, 300 CN) slices and test them on real test sets. The chest x-ray classification model is trained based on the model obtained from LibAUC \cite{yuan2021large}. We report the area under the receiver operating characteristic curve (AUC) from the classification on the real test set. 

\vspace{5pt}
\noindent \textbf{Prompt Generation.}
Since the model performance can be highly dependent on the prompting style, we experiment with different prompting styles: 
\textit{Simple}, where the same prompt template is used for all training images, and \textit{Diverse}, where we randomly select one of the 21 slightly different prompts for each training image. Both styles are used without patient information (\textit{PI}) (e.g., "a brain MRI"), and with \textit{PI} (e.g., "A brain MRI of a \verb|<|age\verb|>| year old \verb|<|sex\verb|>| with \verb|<|diagnosis\verb|>|"). In addition to the general prompt style, we evaluate the model with three different numerical conditioning representations, which correspond to the age here. The three types are: (1) Binned, (2) Word-based, and (3) Numerical. For the \textit{Binned} type, we divide the aging difference based on the value into three groups of "slightly" (\verb|<|5 years), "moderately" (5-10 years), and "significantly" (\verb|>|10 years). The \textit{Word-based} type includes the prompt based on the alphabetical form (e.g., seventy-seven). Finally, the \textit{Numerical} type takes the simplest form of using the number. %
\begin{figure*}[tbh]
    \centering
    \begin{subfigure}{0.49\linewidth}
        \centering
        \includegraphics[width=0.23\linewidth]{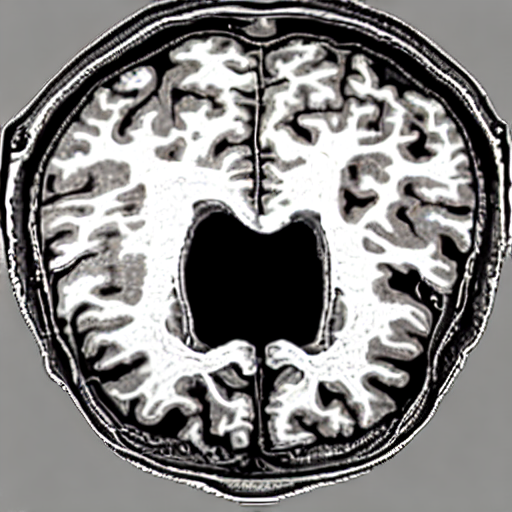}
        \includegraphics[width=0.23\linewidth]{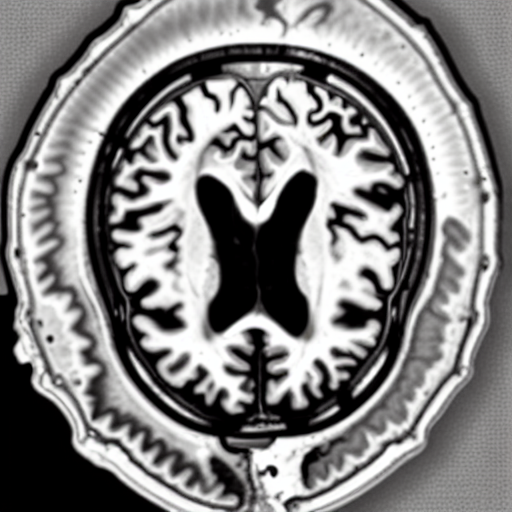}
        \includegraphics[width=0.23\linewidth]{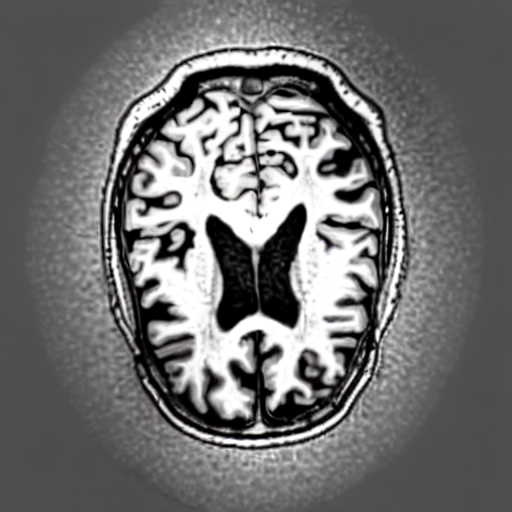}
        \includegraphics[width=0.23\linewidth]{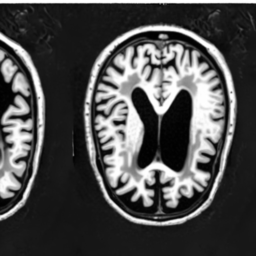}
        \caption{Without \methodname{}}
        \label{fig:no_offset_examples}
    \end{subfigure}
    \begin{subfigure}{0.49\linewidth}
        \centering
        \includegraphics[width=0.23\linewidth]{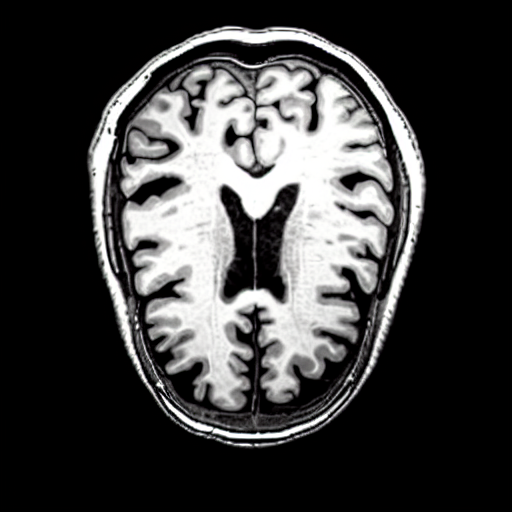}
        \includegraphics[width=0.23\linewidth]{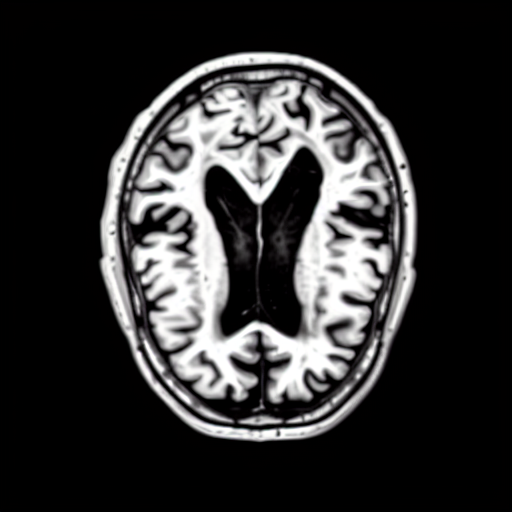}
        \includegraphics[width=0.23\linewidth]{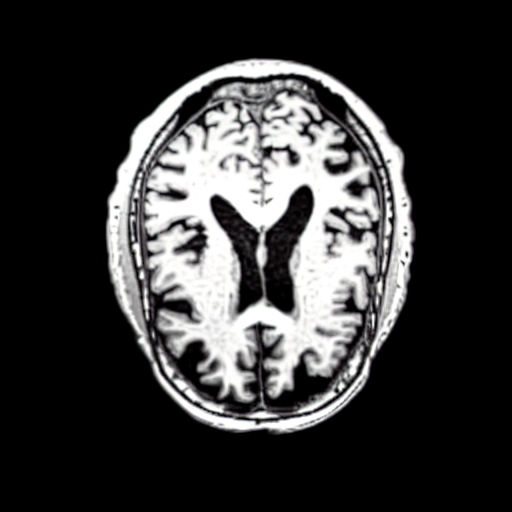}
        \includegraphics[width=0.23\linewidth]{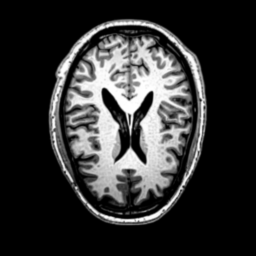}
        \caption{With \methodname{}}
        \label{fig:offset_examples}
    \end{subfigure}
    \caption{Image Generation w. and w/o. \textbf{\method{}} during fine-tuning. Examples generated from left to right using Textual Inversion \cite{gal2022image}, DreamBooth \cite{ruiz2023dreambooth}, Custom Diffusion \cite{kumari2023multi}, and Basic FT \cite{rombach2022high}.}
    \label{fig:noise_offset_comparison}
\end{figure*}

\begin{table}[tb]
    \centering
    \caption{Comparison of different backbones and aging prompt styles for age-conditioned manipulation using InstructPix2Pix + \textbf{\method{}}. \textbf{CD}: Custom Diffusion, \textbf{SD}: Stable Diffusion.}
    \small
    \setlength{\tabcolsep}{3pt}
    \newcommand{\mcw}[1]{\multicolumn{1}{>{\raggedright\arraybackslash}p{#1\linewidth}}}
    \resizebox{\linewidth}{!}{
    \begin{tabular}{@{}
        >{\raggedright\arraybackslash}p{0.18\linewidth}
        >{\raggedright\arraybackslash}p{0.22\linewidth}
        cccc@{}}
    \toprule
    Prompt & Method & FID $\bm{\downarrow}$ & SSIM $\bm{\uparrow}$ & LPIPS $\bm{\downarrow}$ & PSNR $\bm{\uparrow}$ \\ 
    \midrule
    \mcw{0.13}{Binned} 
    & SD \cite{rombach2022high} & 17.05 & \underline{0.74} & \underline{0.13} & \underline{32.78} \\
    & SD + CD \cite{kumari2023multi} & 24.75 & 0.43 & 0.19 & 31.86 \\
    & SD + Basic FT \cite{rombach2022high} & \underline{15.39} & \underline{0.74} & 0.13 & 32.77 \\ 
    \midrule
    \mcw{0.13}{Word} 
    & SD \cite{rombach2022high} & 15.84 & 0.74 & \underline{0.13} & \underline{32.79} \\
    & SD + CD \cite{kumari2023multi} & 27.41 & 0.27 & 0.19 & 30.49 \\
    & SD + Basic FT \cite{kumari2023multi} & \underline{\textbf{15.25}} & \underline{0.75} & \underline{0.13} & 32.78 \\ 
    \midrule
    \mcw{0.13}{Numerical} 
    & SD \cite{rombach2022high} & 16.27 & 0.75 & 0.13 & 32.79 \\
    & SD + CD \cite{kumari2023multi} & 24.05 & 0.32 & 0.23 & 30.70 \\
    & SD + Basic FT \cite{rombach2022high} & \underline{15.37} & \underline{\textbf{0.76}} & \underline{\textbf{0.12}} & \underline{\textbf{32.83}} \\ 
    \bottomrule
    \end{tabular}
    }
    \label{tab:instruct_metrics}
\end{table}

\begin{figure}[tb]
\centering
\begin{tabular}{c@{\hspace{0.1em}}c@{\hspace{0.1em}}c}
     \includegraphics[width=0.24\linewidth]{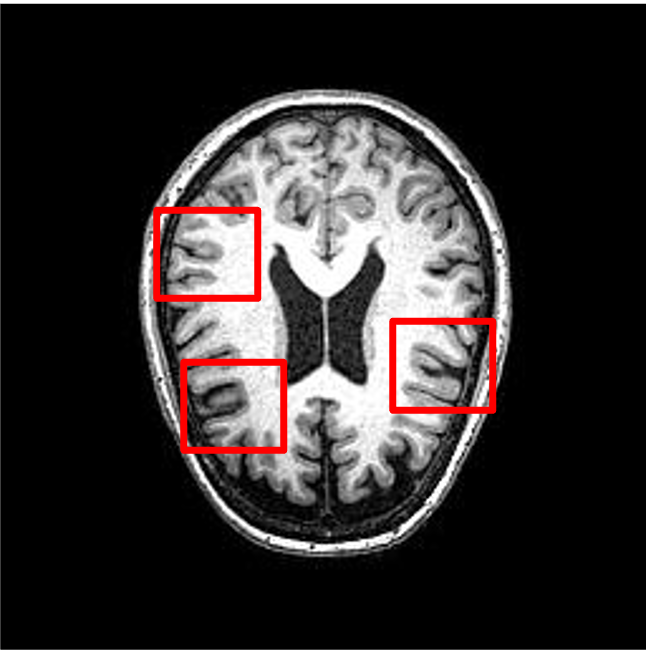} & \includegraphics[width=0.24\linewidth]{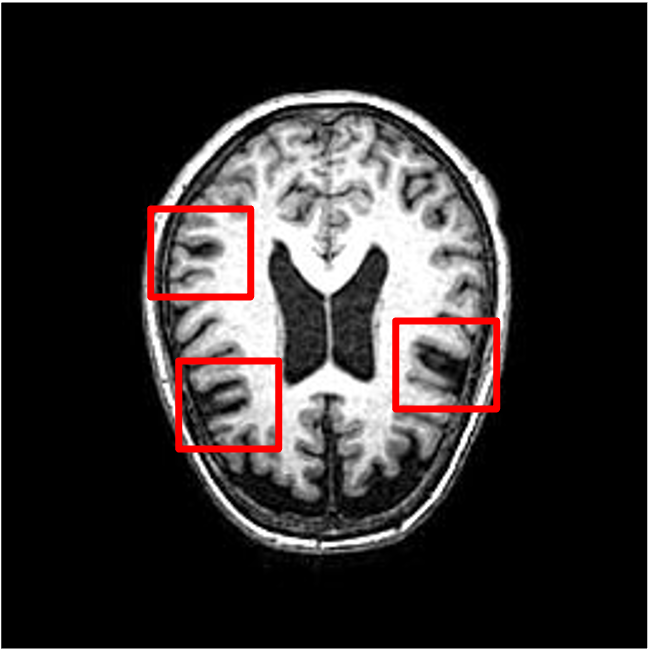} &   \includegraphics[width=0.24\linewidth]{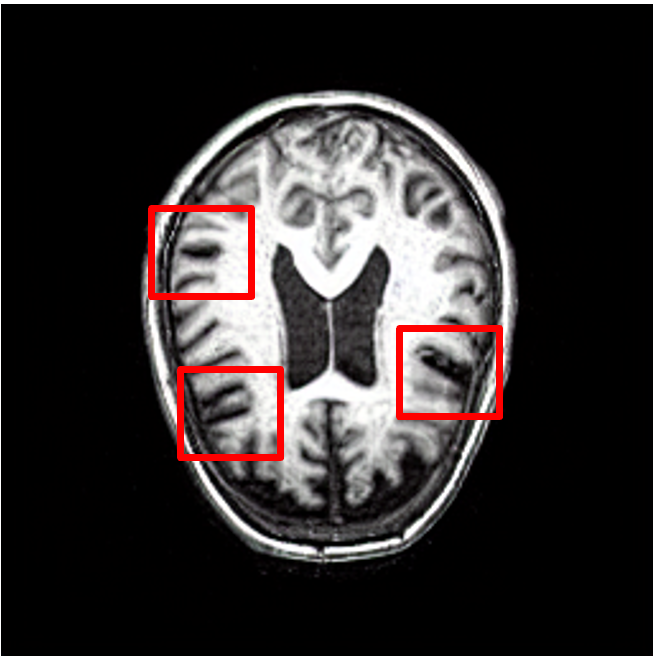}  \\
     Source & Target & Ctf. (\method{}) 
\end{tabular}
	\caption{Brain aging example using the prompt "Age this CN 70 years old female brain MRI into a 77 brain MRI with MCI" via InstructPix2Pix + \method{}. %
 }
	\label{fig:instruct_results}
\end{figure}

\begin{figure}[tb]
    \centering
        \begin{tabular}{c@{\hspace{0.25em}}c@{\hspace{0.25em}}c@{}}

       Original (CN) & Counterfactual (AD) & Diff \\
            \includegraphics[width=0.33\linewidth]{images/Pix2PixZero_examples/cn_orig_0.png}  &  \includegraphics[width=0.33\linewidth]{images/Pix2PixZero_examples/ad_counterfactual_0.png} & \includegraphics[width=0.33\linewidth]{images/Pix2PixZero_examples/diff_ad_counterfactual_0.png} \\
            \includegraphics[width=0.33\linewidth]{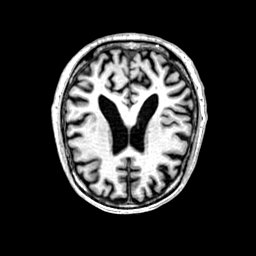}  &  \includegraphics[width=0.33\linewidth]{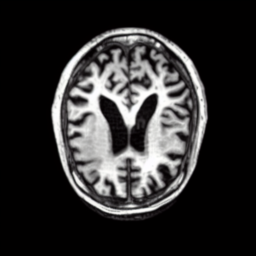} & \includegraphics[width=0.33\linewidth]{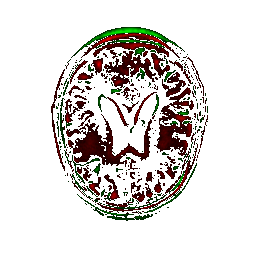} \\
                    \multicolumn{3}{c}{Healthy \quad$\xrightarrow{\hspace*{1.0cm}}$\quad Alzheimer's Disease}\vspace{0.5em} \\

            Original (AD) & Counterfactual (CN) & Diff \\
            \includegraphics[width=0.33\linewidth]{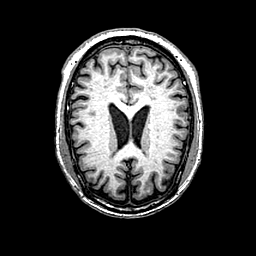}  &  \includegraphics[width=0.33\linewidth]{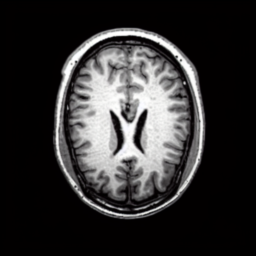} & \includegraphics[width=0.33\linewidth]{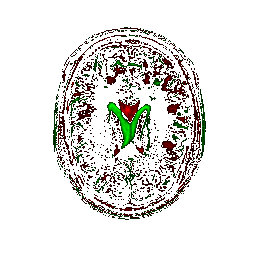} \\
            \includegraphics[width=0.33\linewidth]{images/Pix2PixZero_examples/ad_orig_1.png}  &  \includegraphics[width=0.33\linewidth]{images/Pix2PixZero_examples/cn_counterfactual_1.png} & \includegraphics[width=0.33\linewidth]{images/Pix2PixZero_examples/diff_cn_counterfactual_1.png} \\
                       \multicolumn{3}{c}{Alzheimer's Disease \quad$\xrightarrow{\hspace*{1.0cm}}$\quad Healthy} \\
        \end{tabular}
    \caption{Generated counterfactual MRI slices from Alzheimer's disease to healthy and vice versa using Pix2Pix Zero + \method{}. \raisebox{0.7ex}{\colorbox{red}{}}: Removal, \raisebox{0.7ex}{\colorbox{green}{}}: Addition.}
    \label{fig:pix2pix_counter}
\end{figure}
\vspace{-0.2cm}

\begin{figure*}[tbh]
    \centering
        \begin{tabular}{c@{\hspace{0.1em}}c@{\hspace{0.1em}}c@{\hspace{3em}}c@{\hspace{0.1em}}c@{\hspace{0.1em}}c@{}}
        Orig. & Ctf. & Diff & Orig. & Ctf. & Diff\\
            \includegraphics[width=0.13\linewidth,height=48pt]{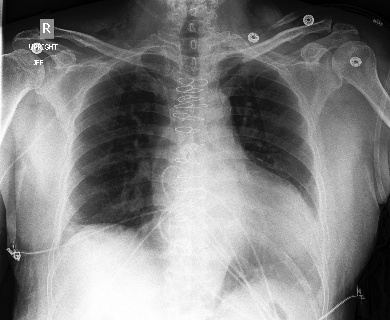}  &  \includegraphics[width=0.13\linewidth,height=48pt]{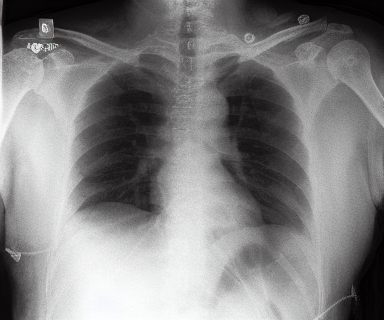} & \includegraphics[width=0.13\linewidth,height=48pt]{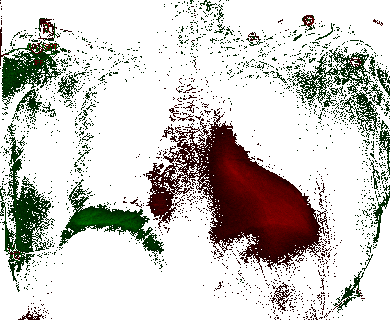} &
            \includegraphics[width=0.13\linewidth,height=48pt] {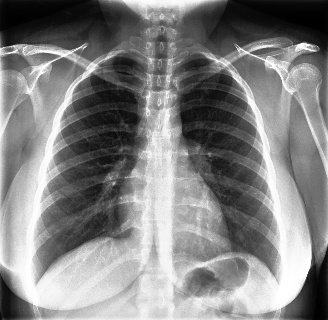}  &  %
            \includegraphics[width=0.13\linewidth,height=48pt]{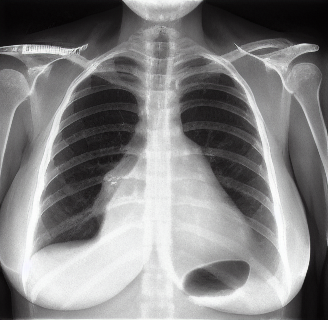} & \includegraphics[width=0.13\linewidth,height=48pt]{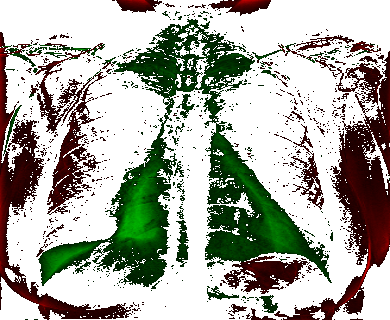} \\
            \multicolumn{3}{c}{
                \makebox[0.39\linewidth][c]{
                    \hspace{-1.9em}Cardiomegaly and Consolidation $\xrightarrow{\hspace*{0.2cm}}$ No finding
                }
            } &
            \multicolumn{3}{c}{
                \makebox[0.39\linewidth][c]{
                    \hspace{0.6em}No finding\quad$\xrightarrow{\hspace*{1.0cm}}$\quad Pleural Effusion
                }
            } \\
            
            \includegraphics[width=0.13\linewidth,height=46pt]{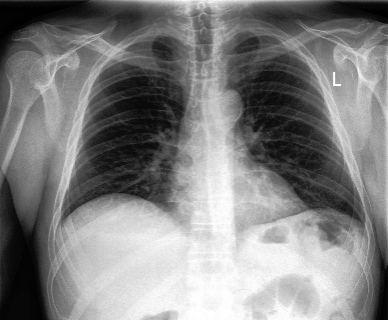}  &  \includegraphics[width=0.13\linewidth,height=46pt]{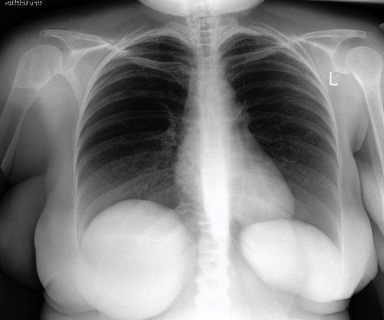} & \includegraphics[width=0.13\linewidth,height=46pt]{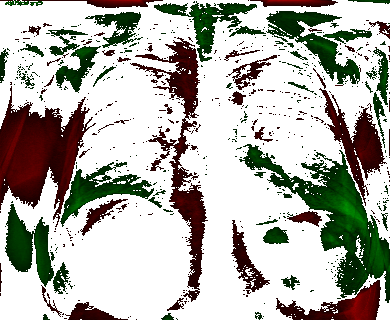} &
            \includegraphics[width=0.13\linewidth,height=46pt]{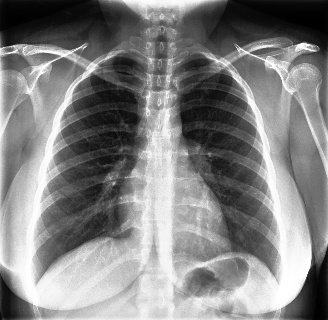}  &  \includegraphics[width=0.13\linewidth,height=46pt]{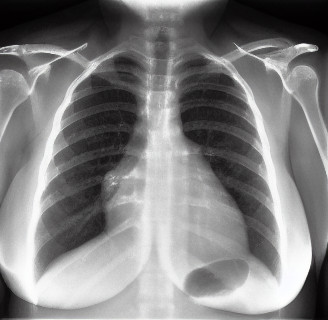} & \includegraphics[width=0.13\linewidth,height=46pt]{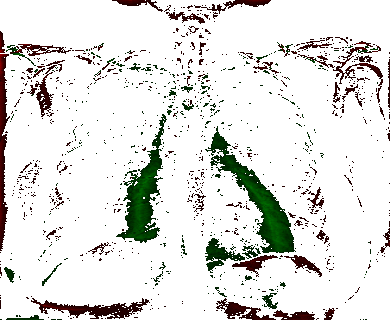} \\
            \multicolumn{3}{c}{
                \makebox[0.39\linewidth][c]{
                 \hspace{-2.5em}Male + No finding\quad$\xrightarrow{\hspace*{0.1cm}}$\quad Female + No finding
                }
            } &
            \multicolumn{3}{c}{
                \makebox[0.39\linewidth][c]{
                    No finding\quad$\xrightarrow{\hspace*{1.0cm}}$\quad Cardiomegaly
                }
            } \\
        \end{tabular}
    \caption{Generated counterfactual samples on CheXpert using Pix2Pix Zero + \method{}. \raisebox{0.7ex}{\colorbox{red}{}}: Removal, \raisebox{0.7ex}{\colorbox{green}{}}: Addition.}
    \label{fig:pix2pix_counter_chest}
\end{figure*}

\subsection{Results}
\begin{table}[tb]
\begin{minipage}[t]{0.49\textwidth}
    \centering
    \caption{Comparison of different prompt styles for text-to-image generation using Stable Diffusion + Basic FT + \method{}. \textbf{PI}: Patient Information.}
    \begin{tabular}{l c c c}
    \hline
        Prompt Style & PI & FID $\bm{\downarrow}$ & KID $\bm{\downarrow}$ \\ \hline
        Simple & \xmark & 60.41 & 0.0475 \\ 
        Simple & \cmark  & \underline{56.49} & \underline{0.0368} \\ \hline
        Diverse & \xmark & 57.99 & 0.0444 \\ 
        Diverse & \cmark & \underline{\textbf{51.35}} & \underline{\textbf{0.0351}} \\ \hline
    \end{tabular}
    \label{tab:unet_prompt_metrics}
\end{minipage}
\hfill
\begin{minipage}[t]{0.49\textwidth}
    \centering
    \caption{Comparison of different prompt styles for text-conditioned image-to-image generation using Pix2Pix Zero + \method{}. \textbf{PI}: Patient Information.}
    
    \begin{tabular}{l c c c}
    \hline
        Prompt Style & PI & FID $\bm{\downarrow}$ & KID $\bm{\downarrow}$ \\ \hline
        Simple & \xmark & 46.79 & 0.0282 \\ 
        Diverse & \xmark& 41.70 & 0.0216 \\ 
        Diverse & \cmark & \textbf{36.10} & \textbf{0.0151} \\ \hline
    \end{tabular}
    \label{tab:pix2pix_prompting}
\end{minipage}
\raggedbottom
\end{table}
\subsubsection{Medical Image Generation}
\label{sec:finetuning}
We present the results of conditional medical image generation with and without our proposed method \method{}, in %
\cref{tab:combined_metrics}, and
\cref{fig:noise_offset_comparison}. In \cref{fig:finetuning_examples}, we show examples of brain MR images generated by the different methods combined with \method{} for two different classes of cognitively normal (CN) and Alzheimer's disease (AD). The results were obtained by fine-tuning the corresponding method on the medical data with \method{}. As shown in \cref{fig:finetuning_examples}, the samples generated through Custom Diffusion tuning are realistically looking while understanding the difference between CN and AD brain properly, which other methods failed at. Textual inversion seems to understand the concept but fails to understand the brain structure properly. Quantitative results on the performance of all our methods are presented in \cref{tab:combined_metrics}.
Qualitative results on the CheXpert \cite{irvin2019chexpert} dataset, along with a user study and ablation of different parameters, optimization algorithm, and more, are included in the supplement.

\vspace{0pt}
\noindent \textbf{Effect of \methodname{}.} \label{sec:noise_offset_results}
Qualitatively, \cref{fig:noise_offset_comparison} shows a significant improvement of the visual realism across all methods when using a drift of $0.1$. The background is consistently black as in real brain MR images; the shape of the brain becomes more realistic, and the white and gray matter structure improves. 
For an analytical evaluation, we calculated the FID between our test data and 200 synthetically generated images from each method (100 CN, 100 AD). The results in \cref{tab:combined_metrics} demonstrate that \method{} improves the ability of the model to generate realistic MRI slices for both healthy brains and brains with Alzheimer's disease. For this reason, all following experiments were done with \method{}.
\subsubsection{Medical Image Manipulation}
The counterfactual conditioning for medical image manipulation is based on pairs of text and images for the two tasks of aging and disease conditioning. We employ the InstructPix2Pix model \cite{brooks2023instructpix2pix} for the aging experiments and the Pix2Pix Zero \cite{parmar2023zero} model for disease conditioning.

\vspace{5pt}
\noindent \textbf{Age-conditioned Manipulation}
The InstructPix2Pix model is based on Stable Diffusion \cite{rombach2022high}. We use three different Stable Diffusion models as the starting point. 1) Stable Diffusion model with basic fine-tuning, 2) Stable Diffusion model fine-tuned using Custom Diffusion \cite{kumari2023multi}, and 3) the original Stable Diffusion model. 
The input to the model includes an image together with an editing prompt, while the output is the edited image (e.g., \cref{fig:instruct_results}). \cref{tab:instruct_metrics} shows the performance of the different starting point models with three prompting styles. As can be seen, the numerical prompts achieve the best overall performance. 
The Custom Diffusion model achieved a much lower performance than the other models, while the model based on the original Stable Diffusion performs slightly worse.

\vspace{5pt}
\noindent \textbf{Disease-conditioned Manipulation}
We use the Pix2Pix Zero model with a basic fine-tuned Stable Diffusion model to generate healthy brain MRIs from ones diagnosed with Alzheimer's Disease and vice versa. 
We generate the counterfactual images by negating the ground truth label of the 200 test samples and conditioning the model on the negated label value and the source image. We compute image quality metrics, as well as the AUC, using a disease classification model trained on 600 real brain MRI slices (300 AD, 300 CN). Additionally, we determine the Structure Similarity Inced (SSIM) between the target and the source image to determine how well the identity of the source image is retained.
The qualitative results in
\cref{fig:pix2pix_counter} illustrate examples from our two editing modes: from AD to CN and from CN to AD, respectively. When transitioning from AD to CN, the model primarily reduces the size of the ventricles. Conversely, in the CN to AD transformation, the ventricle size increases, accompanied by the worsening of brain atrophy.

\subsubsection{Ablation Study}
We provide different ablation studies on the effect of different parameters and components of our model. In the supplementary material, we ablate the effect of $\delta$ and $\tau$ parameters. %

\vspace{5pt}
\noindent \textbf{Prompt Style}
We evaluated four varieties of prompting style, namely \textit{Simple} and \textit{Diverse}, both with and without the patient information (PI) in \cref{tab:unet_prompt_metrics} and \cref{tab:pix2pix_prompting}. In all scenarios, the models fine-tuned with \textit{Diverse} prompts and patient information (PI) performed the best. 

\section{Conclusions}
\label{sec:conclusions}
In this work, we analyzed the generalizability of transferring the knowledge from a pre-trained diffusion model on the natural image domain to another domain with a distribution shift, such as medical imaging. We proposed \methodname{}, an approach for adapting the diffusion model during the fine-tuning process to the target domain by optimizing the latent space. \methodname{} enables the model to generate realistic medical images, while taking advantage of the pretraining domain. We evaluated our method on the counterfactual medical image generation task using text prompts on multiple medical datasets and showed its advantage when combined with SOTA text-to-image and image-to-image generation models.

\section*{Acknowledgments}
This work and collaboration were partially supported by the Munich Center for Machine Learning, BaCaTeC, and the Stanford HAI Hoffman-Yee Award.

{
    \small
    \bibliographystyle{ieeenat_fullname}
    \bibliography{main}
}

\end{document}